\pdfoutput=1

\documentclass[11pt]{article}

\usepackage{booktabs}
\usepackage{multirow}
\usepackage{longtable}
\usepackage{subcaption}
\usepackage{afterpage}

\usepackage[final]{acl}

\usepackage{times}
\usepackage{latexsym}

\usepackage{bm}

\usepackage[T1]{fontenc}

\usepackage[utf8]{inputenc}

\usepackage{microtype}

\usepackage{inconsolata}

\usepackage{graphicx}

\usepackage{tikz}
\usetikzlibrary{positioning,arrows.meta}

\usepackage{pgfplots}
\usepackage{pgfplotstable}
\usetikzlibrary{pgfplots.groupplots}
\pgfplotsset{compat=1.18}

\usepackage[most]{tcolorbox}
\newcommand{\xmark}{\textcolor{red}{\ding{55}}} %
\newcolumntype{C}[1]{>{\centering\arraybackslash}p{#1}}
\newcolumntype{L}[1]{>{\raggedright\arraybackslash}p{#1}}

\usepackage{colortbl}
\usepackage[table]{xcolor}
\usepackage{tabularx}
\usepackage{enumitem}
\setlist[itemize]{itemsep=0pt}
\usepackage{etoolbox}
\usepackage{pgf}
\usepackage{array}
\usepackage{colortbl}

\definecolor{lightred}{HTML}{CC8685}
\definecolor{darkred}{HTML}{FF0000}

\newcommand{\gradientcelld}[8]{
\xdef\lowvalx{#2}%
\xdef\midvalx{#3}%
\xdef\maxvalx{#4}%
\xdef\lowcolx{#5}%
\xdef\midcolx{#6}%
\xdef\highcolx{#7}%
\xdef\opacityx{#8}%
\ifdimcomp{#1pt}{>}{\maxvalx pt}{\cellcolor{\highcolx!100.0!\midcolx!\opacityx}#1}{
\ifdimcomp{#1pt}{<}{\midvalx pt}{%
\ifdimcomp{#1pt}{<}{\lowvalx pt}{\cellcolor{\midcolx!0.0!\lowcolx!\opacityx}#1}{
     \pgfmathparse{int(round(100*(#1/(\midvalx-\lowvalx))-(\lowvalx*(100/(\midvalx-\lowvalx)))))}%
    \xdef\tempa{\pgfmathresult}%
    \cellcolor{\midcolx!\tempa!\lowcolx!\opacityx}#1%
}}{
     \pgfmathparse{int(round(100*(#1/(\maxvalx-\midvalx))-(\midvalx*(100/(\maxvalx-\midvalx)))))}
    \xdef\tempb{\pgfmathresult}%
    \cellcolor{\highcolx!\tempb!\midcolx!\opacityx}#1%
}}
}
\newcommand{\colorcell}[1]{\gradientcelld{#1}{0}{50}{100}{red}{white}{blue}{30}}

\newcommand{\Obl}{\ensuremath{\mathsf{O}}}
\newcommand{\Per}{\ensuremath{\mathsf{P}}}

\newcommand{\corC}{blue!10}
\newcommand{\incorC}{red!10}

\newcommand{\deductionCC}[4]{
\begin{tabular}{p{2ex}l}
\multicolumn{2}{l}{\fbox{\textbf{Label}: \texttt{#1}}} \\
\textsf{P1}:&#2\\
\textsf{P2}:&#3\\ \hline
\textsf{C}:&#4
\end{tabular}
}

\newcommand{\deductionCshort}[3]{%
\begin{tabular}{@{}l@{\hskip 0.5em}p{6.4cm}@{}} %
\textbf{P1}:&#1\\
\textbf{P2}:&#2\\ \hline
\textbf{C}:&#3
\end{tabular}
}

\usepackage{outlines}
\usepackage{float}
\usepackage{comment}

\usepackage{listings}
\usepackage{multicol}

\usepackage{fontawesome5} %

\usepackage{pifont} %

\definecolor{green2}{HTML}{1B9E77}
\newcommand{\greenb}[1]{\textcolor{green2}{#1}}

\definecolor{red2}{HTML}{D55E00}
\newcommand{\redb}[1]{\textcolor{red2}{#1}}

\newcommand{\todo}[1]{\textcolor{black}{#1}}

\newcommand{\ModelSuccessEnt}{\textcolor{green!60!black}{\ding{51}}\ \texttt{valid}}
\newcommand{\ModelFailureEnt}{\xmark\ \texttt{valid}}

\newcommand{\ModelSuccessNonEnt}{\textcolor{green!60!black}{\ding{51}}\ \texttt{invalid}}
\newcommand{\ModelFailureNonEnt}{\xmark\ \texttt{invalid}}

\newcommand{\ErrorExampleBox}[7]{
\begin{tcolorbox}[
    colback=white,      %
    colframe=black!50,  %
    coltext=black,      %
    colbacktitle=gray!15,  %
    coltitle=black,        %
    title=\parbox{\linewidth}{
        \begin{tabular}{@{}ll}
            \textbf{Modal:} \texttt{#1} & \textbf{Inference Pattern:} \texttt{#2} \\
            \textbf{Gold Label:} \texttt{#3} & \textbf{Content Type:} \texttt{#4}
        \end{tabular}
    }
]
\begin{tabular}{@{}L{2ex}L{0.85\linewidth}@{}}
\textbf{P1:} & #5 \\
\textbf{C:}  & #6
\end{tabular}
#7
\end{tcolorbox}
}

\newcommand{\ErrorExampleSylloBox}[8]{
\begin{tcolorbox}[
    colback=white,      %
    colframe=black!50,  %
    coltext=black,      %
    colbacktitle=gray!15,  %
    coltitle=black,        %
    title=\parbox{\linewidth}{
        \begin{tabular}{@{}ll}
            \textbf{Modal:} \texttt{#1} & \textbf{Inference Pattern:} \texttt{#2} \\
            \textbf{Gold Label:} \texttt{#3} & \textbf{Content Type:} \texttt{#4}
        \end{tabular}
    }
]
\begin{tabular}{@{}L{2ex}L{0.85\linewidth}@{}}
\textbf{P1:} & #5 \\
\textbf{P2:} & #6 \\
\textbf{C:}  & #7
\end{tabular}
#8
\end{tcolorbox}
}

\newenvironment{ExampleBox}
  {\begin{tcolorbox}[breakable, colback=white, colframe=black!50, boxrule=0.5pt, arc=2pt, left=6pt, right=6pt, top=4pt, bottom=4pt]}
  {\end{tcolorbox}}

\newcommand{\ModelRes}[6]{
\begin{center}
\begin{tabular}{@{}l C{2cm}@{}}
\toprule
\textbf{Model (#1)} & \textbf{Prediction} \\
\midrule
gpt-4o-mini            & #2 \\
gpt-4o                 & #3 \\
llama-3.1-8b-In  & #4 \\
llama-3.3-70b-In & #5 \\
phi-4                  & #6 \\
\bottomrule
\end{tabular}
\end{center}
}

\definecolor{softred}{rgb}{0.75, 0.15, 0.15}
\newenvironment{revision}
  {\begingroup\color{black}} %
  {\endgroup}              %

\title{Normative Reasoning in Large Language Models: A Comparative Benchmark from Logical and Modal Perspectives}

\author{Kentaro Ozeki\textsuperscript{1,2}, Risako Ando\textsuperscript{1}, Takanobu Morishita\textsuperscript{1}, Hirohiko Abe\textsuperscript{1},\\ \textbf{Koji Mineshima\textsuperscript{1}, Mitsuhiro Okada\textsuperscript{1}}\\
  \textsuperscript{1}Keio University, Tokyo, Japan\\ %
  \textsuperscript{2}University of Tokyo, Tokyo, Japan\\ %
  \texttt{kentaro.ozeki@gmail.com}\quad
  \texttt{\{risakochaan,morishita,hirohiko-abe\}@keio.jp}\\
  \texttt{\{minesima,okada\}@abelard.flet.keio.ac.jp}}

\begin{document}
\maketitle
\begin{abstract}
Normative reasoning is a type of reasoning that involves normative or deontic modality, such as obligation and permission.
While large language models (LLMs) have demonstrated remarkable performance across various reasoning tasks, their ability to handle normative reasoning remains underexplored.
In this paper, we systematically evaluate LLMs' reasoning capabilities in the normative domain from both logical and modal perspectives.
Specifically, to assess how well LLMs reason with normative modals, we make a comparison between their reasoning with normative modals and their reasoning with epistemic modals, which share a common formal structure.
To this end, we introduce a new dataset covering a wide range of formal patterns of reasoning in both normative and epistemic domains, while also incorporating non-formal cognitive factors that influence human reasoning.
Our results indicate that, although LLMs generally adhere to valid reasoning patterns, they exhibit notable inconsistencies in specific types of normative reasoning and display cognitive biases similar to those observed in psychological studies of human reasoning.
These findings highlight challenges in achieving logical consistency in LLMs' normative reasoning and provide insights for enhancing their reliability.
All data and code are released publicly at
\url{https://github.com/kmineshima/NeuBAROCO}.

\end{abstract}

\section{Introduction}
\label{sec:introduction}

Recent research and development of large language models (LLMs) has placed increasing emphasis on their reasoning capabilities, particularly in tasks such as mathematical problem-solving and coding~\cite{mahowald2024517}.
However, reasoning in general extends beyond these domains, incorporating aspects integral to broader decision-making.
One such area is \textit{normative reasoning}, which involves normative or deontic modality, such as obligation, permission, and prohibition (negative obligation).

Normative reasoning can be viewed as a specialized facet of social reasoning, which has recently received substantial attention in LLM research \cite{mondorf2024beyond,mahowald2024517,almeida2024ai}.
Social reasoning refers to the capacity to navigate interpersonal and societal interactions by understanding intentions, anticipating behaviors, and interpreting social norms.
The ability of LLMs to perform normative reasoning is particularly important for the deployment of LLMs in contexts requiring adherence to social, ethical, and legal principles, and has also been linked to challenges in AI alignment \cite{ciabattonietal2023,guan2024deliberative}.

\begin{figure}
\centering
\scalebox{0.84}{
\begin{tabularx}{0.56\linewidth}{X}
\textsf{P}: It is \redb{\textbf{obligatory}} to \mbox{answer} your question.  \\ \hline
\textsf{C}: It is \greenb{\textbf{permitted}} to \mbox{answer} your question. \\
\textbf{pattern}: $\redb{\bm{\Obl}} A \Rightarrow \greenb{\bm{\Per}} A$ \\
\textbf{expected}: \textsf{valid} %
\end{tabularx}
\hspace{1ex}
\begin{tabularx}{0.56\linewidth}{X}
\textsf{P}: It is not \greenb{\textbf{permitted}} to answer your question.  \\ \hline
\textsf{C}: It is not \redb{\textbf{obligatory}} to answer your question.  \\
\textbf{pattern}: $\neg \greenb{\bm{\Per}} A \Rightarrow \neg \redb{\bm{\Obl}}A$ \\
\textbf{expected}: \textsf{valid} %
\end{tabularx}
}
\caption{
The two reasoning patterns are logically related (contrapositive) but LLMs often struggle to make consistent predictions.
$\Obl A$ means ``$A$ is obligatory'' and $\Per A$ means ``$A$ is permitted.''
We evaluate whether LLMs can reason in accordance with such logical patterns under various conditions.
\vspace*{-1em}
}
\label{fig:initial-example}
\end{figure}

Normative reasoning involves formal and contentual aspects.
Prior research on normative reasoning in LLMs has primarily focused on contentual aspects, such as cultural and social factors embedded in the content that influence model behavior~\cite{sheng2021societal,navigli2023biases}.
By contrast, although there is growing interest in the logical or formal reasoning abilities of LLMs~\cite{clark2020soft,bertolazzi-etal-2024-systematic,cheng2025empowering}, the logical aspect of normative reasoning in LLMs has received less attention.
A language model capable of reliable normative reasoning should consistently recognize and apply valid reasoning patterns (Figure~\ref{fig:initial-example}).
This leads to the question of the extent to which LLMs can demonstrate logical consistency in normative reasoning.

\begin{table*}[!th]
    \centering
    \scalebox{0.7}{
    \begin{tabularx}{1.4\textwidth}{@{}llXX@{}}
        \toprule
        \textbf{Label} & \textbf{Form} & \textbf{Premise Example} & \textbf{Hypothesis Example}\\
        \midrule
        \rowcolor{\corC}
        \texttt{NotMu-MiNot} & $\neg \Obl A \Rightarrow \Per \neg A$ &  It is not mandatory to take a shower every day. & It is acceptable not to take a shower every day.
        \\
        \rowcolor{\corC}
        \texttt{NotMi-MuNot} & $\neg \Per A \Rightarrow \Obl \neg A$ &  You are not permitted to litter. & It is mandatory not to litter.
        \\
        \rowcolor{\corC}
        \texttt{MiNot-NotMu} & $\Per \neg A \Rightarrow \neg \Obl A$ &  It is permissible not to attend the party. & There is no obligation to attend the party.
        \\
        \rowcolor{\corC}
        \texttt{Mu-Mi} & $\Obl A \Rightarrow \Per A$ &  You must take care of your health. & You can choose to take care of your health.
        \\
        \rowcolor{\corC}
        \texttt{NotMi-NotMu} & $\neg \Per A \Rightarrow \neg \Obl A$  &  It is not acceptable to lie in court. & It is not the case that you must lie in court.
        \\
        \rowcolor{\incorC} \texttt{NotMu-NotMi} & $\neg \Obl A \Rightarrow \neg \Per A$ &  You are not required to use the internet. & You are not allowed to use the internet.
        \\
        \rowcolor{\incorC} \texttt{MiNot-MuNot} & $\Per \neg A \Rightarrow \Obl \neg A$ &  You are allowed not to drive a car. & You must not drive a car.
        \\
        \rowcolor{\incorC} \texttt{Mi-Mu} & $\Per A \Rightarrow \Obl A$  &  It is permissible to help others. & You are required to help others.
        \\
        \rowcolor{\corC}
        \texttt{FC-Or-Elim} & $\Per (A \vee B) \Rightarrow \Per A$  & You may travel to Japan or France. & You may travel to Japan.
        \\
        \rowcolor{\incorC} \texttt{FC-Or-Intro} & $\Per A \Rightarrow \Per (A \vee B)$ &  You may learn to sing. & You may learn to sing or dance.
        \\
        \rowcolor{\incorC} \texttt{Ross-Or-Intro} & $\Obl A \Rightarrow \Obl (A \vee B)$ &  You must tell the truth. & You must tell the truth or lie.\\
        \bottomrule
    \end{tabularx}
    }%
    \caption{11 patterns of basic deontic logic reasoning. %
    ``$\neg$'' denotes negation (not), ``$\vee$'' denotes disjunction (or), and ``$\phi \Rightarrow \psi$'' represents an inference from the premise $\phi$ to the hypothesis $\psi$.
    \texttt{Mi} and \texttt{Mu} in the label refer to permission and obligation, respectively, while \texttt{NotMi} and \texttt{NotMu} denote negations of permission and obligation.
    \texttt{MiNot} and \texttt{MuNot} denote permission and obligation of negations.
    Those in \colorbox{\corC}{blue} are \textit{valid} patterns, while those in \colorbox{\incorC}{red} are \textit{invalid} patterns.}
    \label{tab:sp-schema}
\end{table*}

In this paper, we systematically evaluate the capabilities of LLMs in normative reasoning, in terms of logical validity and invalidity.
To achieve this, we design reasoning tasks that assess LLMs' consistency across formal reasoning patterns as well as the influence of non-formal factors, extending prior studies to the normative domain in two main directions.
First, to highlight the distinctive characteristics of LLMs in normative reasoning, we compare their performance with another type of modal reasoning, \textit{epistemic reasoning}.
Epistemic reasoning involves logical inference with epistemic modals, which concern a reasoner's knowledge and beliefs, a domain in which LLMs have been shown to exhibit gaps in basic reasoning~\cite{holliday-etal-2024-conditional}.
Second, we compare the LLM reasoning with existing findings from cognitive psychology on human biases in normative reasoning, thereby extending results on LLMs' reasoning biases in general~\cite{ando-etal-2023-evaluating,dasgupta2022language,ozeki2024exploring}.
To support these evaluations, we introduce a new dataset designed for a comparative benchmark of normative and epistemic reasoning in LLMs, while also incorporating non-formal cognitive factors known to influence human reasoning.

\begin{revision}
Our key findings are summarized as follows:
\begin{itemize}
\item Even the best-performing models often show inconsistencies in basic normative reasoning, such as the inference from obligation to permission (illustrated in Figure~\ref{fig:initial-example}, left).

\item The models also manifest human-like biases, such as content effects, in both normative and epistemic reasoning.

\item While it is commonly assumed in the cognitive science literature that normative reasoning is easier than epistemic reasoning for humans, due to domain-specific aspects of cognition (Section~\ref{ssec:deontic-vs-epistemic}), our findings show that language models do not necessarily follow this pattern: their relative performance in the two domains varies across tasks.

\item Reasoning involving negation is particularly challenging for the models, as evidenced by performance on the Syllogistic Task. The presence of negation has a greater impact on reasoning difficulty than the distinction between entailment (valid) and non-entailment (invalid), a contrast previously discussed in the literature~\cite{eisape2023systematic,ozeki2024exploring}.

\end{itemize}
\end{revision}

\section{Patterns of Normative Reasoning}
\label{sec:logical-patterns}

\begin{table*}
\centering
\scalebox{0.79}{
\begin{tabular}{llll}
\cellcolor{\corC}
\deductionCC{Cat-MP}
{All B must C}{A is B}{A must C} &
\cellcolor{\corC}
\deductionCC{Cat-MT}
{All B must C}{A is not required to C}{A is not B} &
\cellcolor{\incorC}
\deductionCC{Cat-AC}
{All B must C}{A must C}{A is B} &
\cellcolor{\incorC}
\deductionCC{Cat-DA}
{All B must C}{A is not B}{A is not required to C}
\medskip \\
\cellcolor{\corC}
\deductionCC{Hyp-MP}
{If A is B, A must C}{A is B}{A must C} &
\cellcolor{\corC}
\deductionCC{Hyp-MT}
{If A is B, A must C}{A is not required to C}{A is not B} &
\cellcolor{\incorC}
\deductionCC{Hyp-AC}
{If A is B, A must C}{A must C}{A is B} &
\cellcolor{\incorC}
\deductionCC{Hyp-DA}
{If A is B, A must C}{A is not B}{A is not required to C}
\end{tabular}
}
\caption{8 patterns of normative (deontic) syllogisms.
\textsf{P1}: Major Premise, \textsf{P2}: Minor Premise, \textsf{C}: Conclusion;
\texttt{Cat}: Categorical,
\texttt{Hyp}: Hypothetical;
\texttt{MP}: Modus Ponens,
\texttt{MT}: Modus Tollens,
\texttt{AC}: Affirming the Consequent,
\texttt{DA}: Denying the Antecedent.
$B$ and $C$ represent predicates, and $A$ represents a term.
Those in \colorbox{\corC}{blue} are \textit{valid} patterns, while those in \colorbox{\incorC}{red} are \textit{invalid} patterns.
}
\label{tab:mp-schema}
\end{table*}

By normative reasoning, we refer to reasoning with deontic modals such as obligation, permission, and prohibition.
\begin{revision}
In this study, we focus on two basic types of normative reasoning:
(1) \textbf{Deontic Logic Reasoning}, which involves single-premise logical inferences and is used to evaluate basic understanding of modal concepts, including challenges specific to modality of obligation, such as Ross's paradox and the free choice inference; and
(2) \textbf{Syllogistic Reasoning}, which involves multi-premise logical inferences and incorporates normative rules and generalizations, including patterns involving universal quantification and conditional statements.
\end{revision}

\subsection{Deontic Logic Reasoning} %
\label{ssec:deontic-logic}

Deontic logic is a formal theory of normative reasoning \cite{von1951deontic,gabbay2013handbook}.
In deontic logic, obligations such as ``It is obligatory that $A$'' and ``One must do $A$''
are symbolized as $\Obl A$, while permissions such as ``It is permissible that $A$'' and ``One may do $A$'' are symbolized as $\Per A$.
Obligation and permission are analyzed as deontic necessity and possibility, where
$\neg \Obl A$ (``It is not obligatory to do $A$'') is equivalent to $\Per \neg A$ (``It is permitted not to do $A$'') and $\neg \Per A$ (``It is not permitted to do $A$'') is equivalent to $\Obl \neg A$ (``It is obligatory not to do $A$'').
Obligation and permission, when combined with negation, follow a traditional inference scheme known as ``The Deontic Square''~\cite{sep-logic-deontic}.

\begin{center}
\begin{tikzpicture}[node distance=4cm, every node/.style={font=\small}]

\node (O) at (0, 1) {Obligatory ($\Obl A$)};
\node (I) at (4, 1) {Impermissible ($\neg \Per A$)};
\node (P) at (0, 0) {Permissible ($\Per A$)};
\node (OM) at (4, 0) {Omissible ($\neg \Obl A$)};

\draw[dashed, red] (O) -- (OM);
\draw[dashed, red] (I) -- (P);

\draw[->] (O) -- (P);
\draw[->] (I) -- (OM);

\end{tikzpicture}
\end{center}

\noindent
For example, an inference from $\Obl A$ (``It is obligatory that $A$'') to $\Per A$ (``It is permissible that $A$'') is \textit{logically valid} in the sense that if the premise is true, the conclusion is also true. This is indicated by the arrow from $\Obl A$ to $\Per A$ in the square.
Similarly, an inference from $\neg \Per A$ (``It is not permissible that $A$'') to $\neg \Obl A$ (``It is not obligatory that $A$'') is also logically valid.
The dotted lines in the square indicate that these pairs are a contradiction.
Table~\ref{tab:sp-schema} presents all formal patterns of deontic logic reasoning examined in the present work.

\begin{revision}
All foregoing inference patterns are valid within Standard Deontic Logic (SDL).
However, it is well known that SDL fails to adequately capture certain intuitively valid or invalid patterns of deontic reasoning.
For instance, consider the inference from ``You may eat an apple or a banana'' to ``You may eat an apple'' ($\Per (A \vee B) \Rightarrow \Per A$).
While this inference is intuitively plausible, it is not valid in SDL, as it relies on disjunction elimination, a rule invalid under the standard interpretation of disjunction: from $A \vee B$ one cannot infer $A$, and SDL inherits this property.
This phenomenon is known as the Free Choice paradox~\cite{kamp1973free}, labeled \texttt{FC-Or-Elim} in Table~\ref{tab:sp-schema}.

Conversely, suppose the inference from ``You must post the letter'' to ``You must post the letter or burn it'' ($\Obl A \Rightarrow \Obl (A \vee B)$). Although intuitively questionable, this inference is valid in SDL, due to the classical logical principle that $A$ entails $A \vee B$. This discrepancy is known as Ross's paradox~\cite{ross1941imperative}, labeled \texttt{Ross-Or-Intro} in Table~\ref{tab:sp-schema}.

These paradoxes reveal a fundamental tension between logical validity in SDL and human judgments in normative reasoning.
In particular, they raise the question of how disjunctions are interpreted in normative contexts:
do models follow logical behavior, in which ``$A \vee B$ does not entail $A$'' and ``$A$ entails $A \vee B$''?
Or do they exhibit modality-specific reasoning, where ``$\Per (A \vee B)$ entails $\Per A$'' (\texttt{FC-Or-Elim}), and ``$\Obl A$ does not entail $\Obl (A \vee B)$'' (\texttt{Ross-Or-Intro}), thereby aligning with intuitive interpretations of normative language?

One of the central goals of our evaluation of deontic reasoning is to test which of these patterns LLMs follow:
whether they behave in accordance with standard logical disjunction, or exhibit reasoning patterns that are characteristic of deontic modality and human intuition.
\end{revision}

\begin{table*}[!th]
    \centering
    \scalebox{0.7}{
    \begin{tabularx}{1.4\textwidth}{@{}llXX@{}}
        \toprule
        \textbf{Label} & \textbf{Form} &  \textbf{Premise} & \textbf{Hypothesis}\\
        \midrule
        \rowcolor{\corC}
        \texttt{NotMu-MiNot} & $\neg \Box A \Rightarrow \Diamond \neg A$ & It is not certain that the economy will recover quickly. & The economy might not recover quickly.
        \\
        \rowcolor{\corC}
        \texttt{NotMi-MuNot} & $\neg \Diamond A \Rightarrow \Box \neg A$  & It is not possible that a person can read a 500-page book in one hour. & It is certain that a person cannot read a 500-page book in one hour.
        \\
        \rowcolor{\corC}
        \texttt{MiNot-NotMu} & $\Diamond \neg A \Rightarrow \neg \Box A$  &  It is possible that the theory will not be proven wrong. & It is not known that the theory will be proven wrong.
        \\
        \rowcolor{\corC}
        \texttt{Mu-Mi} & $\Box A \Rightarrow \Diamond A$ &  It is established that the Earth's climate is changing. & There's a chance that the Earth's climate is changing.
        \\
        \rowcolor{\corC}
        \texttt{NotMi-NotMu} & $\neg \Diamond A \Rightarrow \neg \Box A$  &  It is not possible that a person can perfectly recall every event in their life. & It is not certain that a person can perfectly recall every event in their life.
        \\
        \rowcolor{\incorC} \texttt{NotMu-NotMi} & $\neg \Box A \Rightarrow \neg \Diamond A$ & It is not certain that a person's personality is fixed at birth. & It is not possible that a person's personality is fixed at birth.
        \\
        \rowcolor{\incorC} \texttt{MiNot-MuNot} & $\Diamond \neg A \Rightarrow \Box \neg A$ & It is possible that the painting is not authentic. & It is known that the painting is not authentic.
        \\
        \rowcolor{\incorC} \texttt{Mi-Mu} & $\Diamond A \Rightarrow \Box A$  &  There is a possibility that the universe is teeming with life. & Life must have been teeming in the universe.
        \\
        \bottomrule
    \end{tabularx}
    }%
    \caption{8 patterns of \textit{epistemic} logic reasoning matched to those of normative logic reasoning.
    \texttt{Mi} and \texttt{Mu} in the label refer to epistemic possibility and epistemic necessity, respectively, while \texttt{NotMi} and \texttt{NotMu} denote negations of epistemic possibility and epistemic necessity.
    \texttt{MiNot} and \texttt{MuNot} denote epistemic possibility and epistemic necessity of negations.
    }
    \label{tab:sp-schema-epistemic}
\end{table*}

\subsection{Syllogistic Reasoning}
\label{ssec:modal-syllogism}

\begin{revision}

As a form of inference involving multiple premises, we focus on \textit{syllogisms}, building on prior relevant studies~\cite{ando-etal-2023-evaluating,dasgupta2022language,eisape2023systematic,ozeki2024exploring}.
Syllogism is the type of logical reasoning drawing a conclusion (\textbf{C}) from two premises (\textbf{P1}, \textbf{P2}).
We consider cases in which one of the premises (\textbf{P1}) expresses a normative rule or generalization.
When \textbf{P1} is a universally quantified statement, the inference is called \textit{categorical syllogism} (\texttt{Cat});
when \textbf{P1} takes the form of an ``if-then'' conditional sentence, the inference is called a \textit{hypothetical syllogism} (\texttt{Hyp}).

To examine whether the presence or absence of negation affects the difficulty of inference,
we further classify syllogisms into four distinct patterns: Modus Ponens (\texttt{MP}), Modus Tollens (\texttt{MT}), Affirming the Consequent (\texttt{AC}), and Denying the Antecedent (\texttt{DA}).
The following is an instance of a categorical syllogism with the Modus Tollens pattern (\texttt{Cat-MT}), a logically valid form of inference.

\smallskip

\begin{ExampleBox}
\deductionCshort{All first-year students are required to submit assignments.}
{Mia is not required to submit assignments.}
{Mia is not a first-year student.}
\end{ExampleBox}

\smallskip

As an example of an invalid syllogism, the following is a hypothetical syllogism with Denying the Antecedent pattern (\texttt{Hyp-DA}):

\smallskip

\begin{ExampleBox}
\deductionCshort{If Mia is a first-year student, then she must submit assignments.}
{Mia is not a first-year student.}
{Mia is not required to submit assignments.}
\end{ExampleBox}

\smallskip

\noindent
In these patterns, negation appears in both \textbf{P2} and \textbf{C}, while the inference is logically valid in \texttt{Cat-MT} but invalid in \texttt{Hyp-DA}.
This contrast provides a useful test case for evaluating how the presence of negation and the distinction between valid and invalid reasoning affect the difficulty of inference.

There are a total of eight inference patterns, all of which are presented in Table~\ref{tab:mp-schema}.
Each sentence (involving normative modality) in normative syllogisms is expressed in various expressions, as shown in Table~\ref{tab:normative-template} in Appendix \ref{appendix:dataset}.

\end{revision}

\begin{table*}[t]
\small
\centering
\begin{tabular}{@{}l p{6.3cm} p{6.9cm}@{}}
\toprule
\textbf{Content Type} & \textbf{Description} & \textbf{Example Sentences} \\
\midrule
\textbf{Congruent} &
Premise and conclusion align with common sense. &
Deontic: \textit{It is not obligatory to eat breakfast} \\
&& Epistemic: \textit{It is not certain that AI will replace all jobs} \\
\addlinespace
\textbf{Incongruent} &
Premise or conclusion contradicts common sense. &
Deontic: \textit{It is not obligatory to care for your children} \\
&& Epistemic: \textit{It is not certain that fire needs oxygen} \\
\addlinespace
\textbf{Nonsense} &
Sentences use nonsensical or made-up words. &
Deontic: \textit{It is not obligatory to flibbertigibbet} \\
&& Epistemic: \textit{It is not certain that the flooglehorp grimples the zizzle.} \\
\bottomrule
\end{tabular}
\caption{Three types of content used to analyze content effects in deontic and epistemic reasoning.}
\label{tab:content-types}
\end{table*}

\section{Non-Formal Factors}

To examine how normative reasoning is influenced by factors beyond formal patterns, we consider the effects of modality (domain specificity) and inference content (content effects).

\subsection{Domain Specificity}
\label{ssec:deontic-vs-epistemic}

\noindent
A key question in the study of normative reasoning is whether it exhibits distinctive characteristics that set it apart from other types of reasoning.
An influential view in cognitive science holds that reasoning is domain-specific, meaning that reasoning in different domains is governed by distinct cognitive mechanisms.
This is known as \textit{domain specificity} of reasoning
\cite{cosmides1989logic,cosmides-Tooby1992,fiddick2004domains}.
\citet{seals-shalin-2024-evaluating} report that LLMs perform better in reasoning with social content than with non-social content, mirroring human reasoning, though the effect is less prominent in LLMs than in humans.
However, it remains unclear whether this domain-specific reasoning ability extends to normative reasoning as in the case of humans~\cite{cheng1985pragmatic}.

We investigate this question by comparing normative reasoning with \textit{epistemic reasoning}.
Similar to deontic logic for normative reasoning, the formal framework developed to model epistemic reasoning is known as \textit{epistemic logic} \cite{Hintikka1962,sep-logic-epistemic}. Epistemic logic includes two types of modal expressions reflecting necessity and possibility in terms of knowledge and beliefs.
\textbf{Epistemic necessity} indicates that something is certain given the evidence, and includes expressions such as \textit{It is known that...}, \textit{It is certain that...}, or the modal \textit{must} (e.g., \textit{She must be an expert}).
\textbf{Epistemic possibility} indicates that something may be true given the reasoner’s knowledge and beliefs, and includes expressions such as \textit{It is possible that...} or the modal \textit{might} (e.g., \textit{She might be an expert}).

Some inference patterns that are valid in deontic logic are also valid in epistemic logic, and the same holds for invalid patterns.
For example, the following is an epistemic version of an invalid hypothetical syllogism with Denying the Antecedent pattern (\texttt{Hyp-DA}), whose deontic counterpart we saw in Section \ref{ssec:modal-syllogism}.

\smallskip

\begin{ExampleBox}
\deductionCshort{If a student misses orientation, then they must be new to the program.}
{Sam did not miss orientation.}
{Sam must not be new to the program. }
\end{ExampleBox}

\smallskip

All patterns of epistemic logic reasoning and epistemic syllogisms examined in this study are listed in Table~\ref{tab:sp-schema-epistemic} and in Table~\ref{tab:mp-schema-epistemic} in Appendix~\ref{appendix:dataset}, respectively.
Variations in the expressions used to represent epistemic modality are shown in Table~\ref{tab:epistemic-template} in Appendix~\ref{appendix:dataset}.

In the literature, \citet{holliday-etal-2024-conditional} provide a systematic evaluation of LLMs' reasoning with epistemic modality.
The present study extends this line of work by investigating how the modality of reasoning---deontic versus epistemic---affects LLM behavior, through a comparative analysis using the two tasks described above.

\subsection{Content Effects}
\label{ssec:content-types}

\begin{revision}

The \textit{content effect} is the human tendency to accept inferences whose conclusions align with one’s beliefs and to reject those whose conclusions do not, regardless of their logical validity~\cite{evans1993human}.
Recent studies have shown that LLMs exhibit similar behavior:
they tend to make reasoning errors when conclusions contradict common-sense beliefs~\cite{ando-etal-2023-evaluating,dasgupta2022language,ozeki2024exploring}.

To analyze content effects in deontic reasoning, we categorize each inference into one of three content types based on the nature of its premise and conclusion.
These categories distinguish whether the statements are consistent with common-sense knowledge, contradict it, or are nonsensical.
The three types are summarized in Table~\ref{tab:content-types}.

\end{revision}

\section{Experiments}

In this section, we describe the data and task formats, the models used in the experiments, and the evaluation settings.

\subsection{Data and Task Formats}

We evaluate two types of inference problems, \textbf{Deontic Logic} and \textbf{Syllogistic}, as described in Section \ref{sec:logical-patterns}.
Each type is divided into normative reasoning and epistemic reasoning problems using various modal vocabularies as explained in Section~\ref{ssec:deontic-vs-epistemic}.
All problems are written in English.

For the {Deontic Logic} task, we manually created 11 templates for normative inference (see Table~\ref{tab:sp-schema}) and 8 templates for epistemic inferences (see Table~\ref{tab:sp-schema-epistemic}).
For the {Syllogistic} task, we created 8 templates for normative and epistemic inferences (see Table~\ref{tab:mp-schema}).
We then instantiated these templates with 20 concrete words for each of the three content types explained in Section~\ref{ssec:content-types}, using Gemini 1.5 Pro, which was not used in the evaluation.
Finally, we manually refined the resulting instances, when necessary, to ensure consistency and quality.

The {Deontic Logic} task includes 640 problems for normative reasoning (360 valid, 280 invalid) and 480 problems for epistemic reasoning (300 valid, 180 invalid).
The {Syllogistic} task includes 480 problems (240 valid and 240 invalid) for normative and epistemic reasoning, respectively.

\subsection{Models}

For our analysis, we evaluated 5 recently released language models, using their instruction-tuned versions when available.
GPT-4o and GPT-4o-mini~\cite{openai2024gpt4o} are state-of-the-art closed-weight models, accessed via the OpenAI API.
Llama-3.1-8B-In (8B parameters) and Llama-3.3-70B-In (70B parameters) are instruction-tuned versions of open-weight Llama 3 models~\cite{dubey2024llama3}.
Phi-4 is a 14B-parameter open-weight model that achieves strong reasoning performance relative to its size~\cite{abdin2024phi4}.

\subsection{Evaluation}

The models are evaluated based on the accuracy of their predictions against the expected answers.
We conduct experiments using three types of prompts.
In the Zero-Shot setting, the prompt contains only the instruction for the task and the problem.
In the Few-Shot setting, exemplars for in-context learning are included~\cite{brown2020language}.
Here, we provide a single exemplar with expected answers for each reasoning pattern, where the patterns differ depending on whether the problem belongs to the Normative or Epistemic domain.
In the Chain-of-Thought (CoT) setting, the model is prompted to generate a sequence of reasoning steps that lead to the final conclusion~\cite{wei2022chain}.
Specifically, we adopt the Zero-Shot CoT approach~\cite{kojima2022large}.

The evaluation is performed in a single run with the temperature set to 0.0 to make the model responses deterministic.
When the expected output is a single word, we limit the maximum output tokens to 10.
In the CoT setting, the maximum output token is extended to 1024.
Other hyperparameters are kept at their default values.

\begin{table*}[!t]
\centering
\scalebox{0.73}{%
\begin{tabular}{l
  *{3}{c}@{\quad}*{3}{c}@{\quad}
  *{3}{c}@{\quad}*{3}{c}}
\toprule
\multirow{3}{*}{\textbf{Model}} &
  \multicolumn{6}{c}{\textbf{Deontic Logic}} &
  \multicolumn{6}{c}{\textbf{Syllogistic}} \\
\cmidrule(lr){2-7} \cmidrule(lr){8-13}
 & \multicolumn{3}{c}{\textbf{Normative}} & \multicolumn{3}{c}{\textbf{Epistemic}}
 & \multicolumn{3}{c}{\textbf{Normative}} & \multicolumn{3}{c}{\textbf{Epistemic}} \\
\cmidrule(lr){2-4} \cmidrule(lr){5-7} \cmidrule(lr){8-10} \cmidrule(lr){11-13}
 & \textbf{Zero} & \textbf{Few} & \textbf{CoT}
 & \textbf{Zero} & \textbf{Few} & \textbf{CoT}
 & \textbf{Zero} & \textbf{Few} & \textbf{CoT}
 & \textbf{Zero} & \textbf{Few} & \textbf{CoT} \\
\midrule
gpt-4o-mini      & \colorcell{81.41} & \colorcell{87.19} & \colorcell{80.47}
                  & \colorcell{84.79} & \colorcell{85.21} & \colorcell{82.08}
                  & \colorcell{69.58} & \colorcell{82.08} & \colorcell{76.91}
                  & \colorcell{63.33} & \colorcell{72.29} & \colorcell{76.56} \\
gpt-4o           & \colorcell{84.22} & \colorcell{97.03} & \colorcell{85.31}
                  & \colorcell{93.75} & \colorcell{92.29} & \colorcell{91.67}
                  & \colorcell{78.33} & \colorcell{92.29} & \colorcell{85.90}
                  & \colorcell{62.92} & \colorcell{83.96} & \colorcell{76.88} \\
llama-3.1-8B-In  & \colorcell{70.31} & \colorcell{73.28} & \colorcell{73.28}
                  & \colorcell{76.04} & \colorcell{86.83} & \colorcell{67.29}
                  & \colorcell{56.25} & \colorcell{46.67} & \colorcell{58.33}
                  & \colorcell{57.08} & \colorcell{52.50} & \colorcell{54.79} \\
llama-3.3-70B-In & \colorcell{78.75} & \colorcell{94.06} & \colorcell{80.78}
                  & \colorcell{91.67} & \colorcell{91.25} & \colorcell{95.63}
                  & \colorcell{56.25} & \colorcell{46.67} & \colorcell{58.33}
                  & \colorcell{57.08} & \colorcell{52.50} & \colorcell{54.79} \\
phi-4            & \colorcell{82.50} & \colorcell{80.94} & \colorcell{78.59}
                  & \colorcell{92.08} & \colorcell{84.79} & \colorcell{88.33}
                  & \colorcell{75.62} & \colorcell{67.29} & \colorcell{80.83}
                  & \colorcell{75.21} & \colorcell{76.25} & \colorcell{66.25} \\
\bottomrule
\end{tabular}%
}
\caption{
Overall accuracy (\%) for Normative and Epistemic problems across the tasks. Prompting strategies: \textbf{Zero} = Zero-Shot, \textbf{Few} = Few-Shot, \textbf{CoT} = Chain-of-Thought.
Shading follows a gradient from \colorbox{red!30}{red} (0\%) to \colorbox{blue!30}{blue} (100\%), with white representing the midpoint (50\%).
}
\label{tab:results-overall}
\end{table*}

\section{Results and Analysis}

\noindent
In this section, we present the results of the normative reasoning tasks. %
Table~\ref{tab:results-overall} presents the overall performance of the models on the Normative problems in the two tasks, in comparison to those on the Epistemic problems.
GPT-4o in the Few-Shot setting achieves the highest accuracy among the models in the Normative domain on all tasks.
Among the smaller models, Phi-4 shows competitive performance to the larger models.

\subsection{Comparison by Prompt Type}
The performance of the models is generally higher in the Few-Shot setting compared to the Zero-Shot setting.
However, Llama models show a decrease in performance with Few-Shot prompts on the Syllogistic task.
Chain-of-Thought prompting contributes to a small improvement or has negative impact in the Deontic Logic task, while it has \todo{a larger impact} on the Syllogistic task.

\subsection{Consistency across Reasoning Patterns}

\noindent
First, to investigate the consistency of the models across reasoning patterns of different forms, we compare the performance of the models on the different normative reasoning tasks.
The complete results and error examples in the Deontic Logic and Syllogistic tasks are presented in Appendix~\ref{appendix:experiments}.

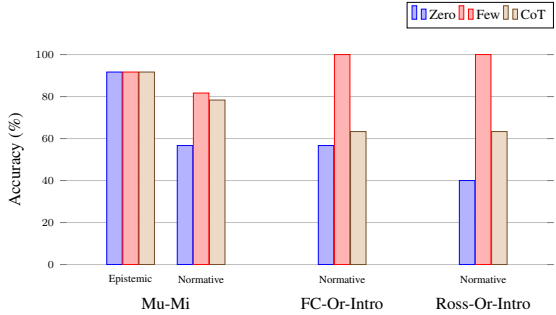
\begin{figure}[!t]
\centering
\scalebox{0.58}{
\begin{tikzpicture}
\begin{axis}[
  xticklabels={Epistemic, Normative, Normative, Normative},
  xtick=data,
  ylabel={Accuracy (\%)},
  major x tick style=transparent,
  ybar=\pgflinewidth,
  ymin=0,
  ymax=100,
  x axis line style={opacity=0},
  x tick label style={rotate=0,anchor=center},
  xticklabel style={yshift=-2mm},
  xtick=data,
  tick label style={font=\scriptsize},
  ymajorgrids=true,
  scale only axis,
  point meta=explicit symbolic,
  enlarge x limits = {abs=1},
  legend columns=-1,
  legend pos=north east,
  legend style={font=\small,at={(1,1.25)}},
  height=0.3\textwidth,
  width=0.7\textwidth,
  extra x ticks={1.5, 4, 6},
  extra x tick labels={Mu-Mi,FC-Or-Intro,Ross-Or-Intro},
  extra x tick style={yshift=-15pt,
  tick label style={rotate=0,font=\normalsize}},
]

\addplot  coordinates {
  (1,91.67)
  (2,56.67)
  (4,56.67)
  (6,40.0)
  };

\addplot  coordinates {
  (1,91.67)
  (2,81.67)
  (4,100.0)
  (6,100.0)
  };

\addplot  coordinates {
  (1,91.67)
  (2,78.33)
  (4,63.33)
  (6,63.33)
  };

\legend{Zero,Few,CoT}
\end{axis}
\end{tikzpicture}
}%

\caption{%
Accuracy (\%) of the best-performing model (GPT-4o) for Mu-Mi, FC-Or-Intro, Ross-Or-Intro patterns in the Deontic Logic task.
The FC-Or-Intro and Ross-Or-Intro patterns are Normative problems only.
}
\label{fig:sp-normative}
\end{figure}

\paragraph{Deontic Logic Task.}
Figure~\ref{fig:sp-normative} illustrates the performance of the best-performing model (GPT-4o) on the most challenging patterns in the Deontic Logic task: \texttt{Mu-Mi} (the inference from obligation to permission) and the paradox-related inferences (\texttt{FC-Or-Intro} and \texttt{Ross-Or-Intro}).
Overall, the models perform well when the patterns align with Standard Deontic Logic.
However, in the \texttt{Mu-Mi} pattern, most models exhibit lower performance, with Llama-3.3-70B-In being the only model to perform well.
For the controversial patterns, namely \texttt{FC-Or-Elim}, \texttt{FC-Or-Intro}, and \texttt{Ross-Or-Intro}, the models demonstrate mixed performance.
They perform well on the \texttt{FC-Or-Elim}, where valid is the expected answer that aligns with common-sense reasoning.
In contrast, LLMs tend to accept the validity of \texttt{FC-Or-Intro} and \texttt{Ross-Or-Intro}, which are expected to be invalid.
Regarding the relative difficulty of valid and invalid patterns, no clear trend is observed.

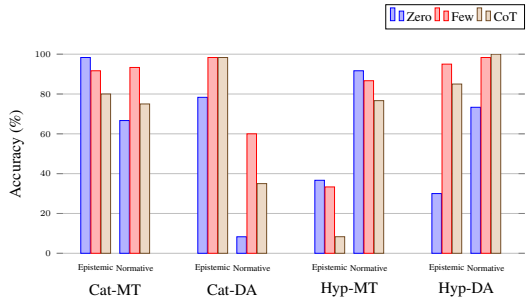
\begin{figure}[!t]
    \centering
    \scalebox{0.55}{
    \begin{tikzpicture}
    \begin{axis}[
      xticklabels={Epistemic, Normative, Epistemic, Normative, Epistemic, Normative,Epistemic, Normative},
      xtick=data,
      ylabel={Accuracy (\%)},
      major x tick style=transparent,
      ybar=\pgflinewidth,
      ymin=0,
      ymax=100,
      x axis line style={opacity=0},
      x tick label style={rotate=0, anchor=center},
      xticklabel style={yshift=-2mm},
      xtick=data,
      tick label style={font=\tiny},
      ymajorgrids=true,
      scale only axis,
      point meta=explicit symbolic,
      enlarge x limits = {abs=1},
      legend columns=-1,
      legend pos=north east,
      legend style={font=\small,at={(1,1.25)}},
      height=0.3\textwidth,
      width=0.7\textwidth,
      extra x ticks={1.5, 4.5, 7.5, 10.5},
      extra x tick labels={Cat-MT,Cat-DA,Hyp-MT,Hyp-DA},
      extra x tick style={yshift=-15pt,
      tick label style={font=\normalsize}},
      bar width=6.5pt,
    ]

    \addplot  coordinates {
      (1,98.33)
      (2,66.67)
      (4,78.33)
      (5,8.3)
      (7,36.67)
      (8,91.67)
      (10,30.0)
      (11,73.33)
      };

    \addplot  coordinates {
      (1,91.67)
      (2,93.33)
      (4,98.33)
      (5,60.0)
      (7,33.33)
      (8,86.67)
      (10,95.0)
      (11,98.33)
      };

    \addplot  coordinates {
      (1,80.0)
      (2,75.0)
      (4,98.33)
      (5,35.0)
      (7,8.33)
      (8,76.67)
      (10,85.0)
      (11,100.0)
      };

    \legend{Zero,Few,CoT}
    \end{axis}
    \end{tikzpicture}
    }%
    \caption{Accuracy (\%) of the best-performing model (GPT-4o) for Cat-MT, Cat-DA, Hyp-MT, Hyp-DA patterns in the Syllogistic task.}
    \label{fig:mp-normative}
\end{figure}

\paragraph{Syllogistic Task.}
Figure~\ref{fig:mp-normative} presents the performance of the best-performing model (GPT-4o) in the Syllogistic task.
Both in the categorical (\texttt{Cat}) and hypothetical (\texttt{Hyp}) cases, the models perform well on the \texttt{MP} and \texttt{AC} patterns, but show lower performance on the \texttt{MT} and in particular \texttt{DA} patterns.
The latter two involve negation in the premises and hypotheses, whereas the former two do not.
Previous studies have shown that LLMs struggle with reasoning involving negation~\cite{truong-etal-2023-language,garcia-ferrero-etal-2023-dataset}.
Our results in the Syllogistic task corroborate this observation.

\subsection{Non-Formal Factors}

\paragraph{Domain Specificity.}
To assess whether models perform better on normative or epistemic reasoning, we compare their performance across both domains.
The overall results for Normative and Epistemic domains are presented in Table~\ref{tab:results-overall}.
In the Syllogistic task, models perform comparably or better on normative problems than on epistemic ones under the Zero-Shot setting, which aligns with human tendency studied in cognitive science.
In contrast, in the Deontic Logic task, normative problems are more challenging than epistemic problems under the Zero-Shot setting.
These results suggest that the relationship between reasoning domain and model performance may vary across tasks and reasoning patterns.

\begin{table}[t]
\centering
\begin{subtable}{\linewidth}
\centering
\caption{Deontic Logic}
\scalebox{0.7}{
\begin{tabular}{lccc}
\toprule
\textbf{Model} & \textbf{Incong.} & \textbf{Cong.} & \textbf{Nonsense} \\
\midrule
gpt-4o-mini & \colorcell{79.09} & \colorcell{84.0} & \colorcell{81.36} \\
gpt-4o & \colorcell{72.27} & \colorcell{94.0} & \colorcell{87.27} \\
llama-3.1-8B-In & \colorcell{72.73} & \colorcell{79.0} & \colorcell{60.0} \\
llama-3.3-70B-In & \colorcell{76.36} & \colorcell{85.50} & \colorcell{75.0} \\
phi-4 & \colorcell{88.0} & \colorcell{75.91} & \colorcell{84.09} \\
\bottomrule
\end{tabular}
}%
\end{subtable}

\medskip

\begin{subtable}{\linewidth}
\centering
\caption{Syllogistic}
\scalebox{0.7}{
\begin{tabular}{lccc}
\toprule
\textbf{Model} & \textbf{Incong.} & \textbf{Cong.} & \textbf{Nonsense} \\
\midrule
gpt-4o-mini & \colorcell{66.25} & \colorcell{75.0} & \colorcell{67.50} \\
gpt-4o & \colorcell{78.75} & \colorcell{83.13} & \colorcell{73.12} \\
llama-3.1-8B-In & \colorcell{66.87} & \colorcell{60.62} & \colorcell{41.25} \\
llama-3.3-70B-In & \colorcell{71.25} & \colorcell{71.25} & \colorcell{70.63} \\
phi-4 & \colorcell{68.75} & \colorcell{79.37} & \colorcell{78.75} \\
\bottomrule
\end{tabular}
}%
\end{subtable}

\caption{Accuracy (\%) per content type for Normative problems on each task (Zero-Shot). \textbf{Incong.} = Incongruent, \textbf{Cong.} = Congruent.}
\label{tab:normative-contents}
\end{table}

\paragraph{Content Effect.}

We observed content effects in both normative and epistemic problems.
Table~\ref{tab:normative-contents} presents the performance of the models on the normative problems of the different content types in each task.
In the Deontic Logic task, most of the models except Phi-4 perform best on problems with congruent content, followed by those with nonsense and incongruent content.
Phi-4 shows the highest performance on incongruent content, followed by nonsense and congruent content.
In the Syllogistic task, problems with nonsense content tend to be more challenging; however, the performance gap narrows among the better-performing models.

\section{Discussion}

\begin{figure*}[t]
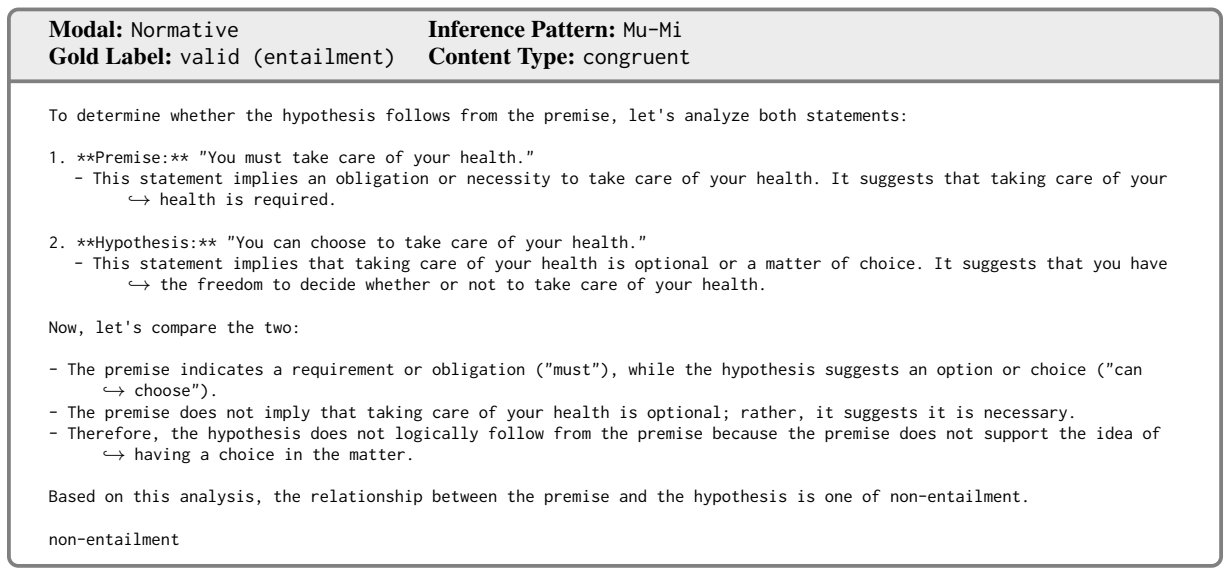


\centering

\begin{tcolorbox}[
    width=\textwidth,         %
    enlarge left by=0mm,      %
    enlarge right by=0mm,
    colback=white,      %
    colframe=black!50,  %
    coltext=black,      %
    colbacktitle=gray!15,  %
    coltitle=black,        %
    bottom=-4pt,
    title=\parbox{\textwidth}{
        \small
        \begin{tabular}{@{}ll}
            \textbf{Modal:} \texttt{Normative} & \textbf{Inference Pattern:} \texttt{Mu-Mi} \\
            \textbf{Gold Label:} \texttt{valid (entailment)} & \textbf{Content Type:} \texttt{congruent}
        \end{tabular}
    }
]
\vspace*{-6pt}
{\scriptsize
\begin{lstlisting}
To determine whether the hypothesis follows from the premise, let's analyze both statements:

1. **Premise:** "You must take care of your health."
   - This statement implies an obligation or necessity to take care of your health. It suggests that taking care of your health is required.

2. **Hypothesis:** "You can choose to take care of your health."
   - This statement implies that taking care of your health is optional or a matter of choice. It suggests that you have the freedom to decide whether or not to take care of your health.

Now, let's compare the two:

- The premise indicates a requirement or obligation ("must"), while the hypothesis suggests an option or choice ("can choose").
- The premise does not imply that taking care of your health is optional; rather, it suggests it is necessary.
- Therefore, the hypothesis does not logically follow from the premise because the premise does not support the idea of having a choice in the matter.

Based on this analysis, the relationship between the premise and the hypothesis is one of non-entailment.

non-entailment
\end{lstlisting}
} %
\end{tcolorbox}
\caption{An example output of GPT-4o with a CoT prompt for reasoning from obligation to permission (\texttt{Mu-Mi}), which includes the expression ``can choose to.''
\vspace*{-1em}
}
\label{fig:CoTexample}
\end{figure*}

\noindent
Our findings highlight several areas for improving the logical consistency of LLMs' normative reasoning.

First, we observed that performance of the LLMs varies not only across different reasoning tasks but also within the same task, depending on the specific reasoning pattern.
Our results on the Deontic Logic task reveal that the reasoning from obligation to permission (\texttt{Mu-Mi}) did not align with expected validity despite its simple form.
This can lead to logical inconsistency, as LLMs tend to accept its logically equivalent contrapositive (\texttt{NotMi-NotMu}) as valid reasoning.
Analysis of Chain-of-Thought outputs suggests that this inconsistency is often rooted in how permission is expressed in the problem.
For instance, GPT-4o did not infer ``You can choose to take care of your health'' from ``You must take care of your health'', interpreting ``can choose to'' as an option rather than a statement of permission \todo{(see Figure~\ref{fig:CoTexample})}.
Our evaluation of syllogistic reasoning further indicates that negation affects the logical consistency of normative reasoning in LLMs, with reasoning patterns involving negation proving more difficult for the models.
At the same time, the results from the Deontic Logic task suggest that the difficulty associated with negation is not solely due to its presence but may also depend on the complexity of the reasoning.

Second, we identified several biases and effects in normative reasoning of LLMs that resemble those observed in human reasoning.
As with other kinds of reasoning in LLMs, normative reasoning is influenced by content effects, which affect the logical consistency of the models.
We also observed domain specificity of normative reasoning.
Our comparison of normative and epistemic reasoning indicates that whether the models perform better in the normative domain than in other domains depends on the specific task.
This finding suggests that conclusions about model performance within a domain should take potential task-specific variations into account.

Throughout our experiments, we compared different prompting strategies, namely Zero-Shot, Few-Shot, and Chain-of-Thought, to examine their impact on the logical consistency of LLMs in normative reasoning.
Our findings indicate that the effectiveness of these strategies varies across tasks and models.
For Deontic Logic and Syllogistic reasoning tasks, Few-Shot prompting generally led to performance improvements, as expected.
This can be attributed to the straightforward nature of leveraging similarities in syntactic structures within reasoning patterns.
However, the observed improvements may result from superficial pattern matching rather than genuine reasoning.
On the other hand, Chain-of-Thought prompting generally produced minimal improvement or even negative effects, a pattern also observed in epistemic reasoning.
A common issue was the introduction of errors in intermediate reasoning steps, which led to incorrect final conclusions.
These observations suggest that Chain-of-Thought does not necessarily enhance robustness in normative reasoning and may instead introduce additional points of failure.

\section{Conclusion}
\label{sec:conclusion}

In this paper, we systematically evaluated the normative reasoning capabilities of LLMs, examining both formal logical structures and cognitive influences.
Our findings emphasize the significance of \textit{formal} validity and invalidity in normative reasoning, extending beyond normative \textit{content}.
While models often align with valid reasoning patterns, they exhibit inconsistencies in specific inferences and cognitive biases similar to human reasoning.
These results highlight challenges in maintaining formal consistency and underscore the need for further improvements to enhance the reliability of normative reasoning in LLMs.

\section*{Limitations}

While our study provides a comprehensive analysis of normative reasoning in LLMs, several limitations should be acknowledged.
First, the nature of normative reasoning and deontic logic remains an open research question.
Our evaluation relies on a specific set of valid reasoning patterns based on the literature on logical and cognitive studies of normative reasoning, but alternative frameworks could yield different insights into LLM performance.
Our study focuses on specific, controlled reasoning tasks rather than open-ended normative deliberation.
While our dataset captures key logical patterns, real-world normative reasoning often involves additional complexities, such as contextual interpretation, ethical considerations, and pragmatic constraints.
Future work could explore how LLMs perform in more applied settings in broader contexts.

Second, the LLMs are continuously evolving, and our findings may not generalize to future models with improved reasoning mechanisms.
The performance of models is affected by updates in training data, architecture, and fine-tuning strategies.
In addition, we compare closed and open-source models, but our experiments rely on API access for proprietary models, limiting our ability to analyze their internal mechanisms.
Differences in fine-tuning, prompt engineering, and instruction-following abilities may also contribute to performance variations, making it challenging to isolate the effects of model architecture alone.

\section*{Acknowledgements}

\todo{We thank the anonymous reviewers for their comments and suggestions.
This work is partially supported by JST CREST Grant Number JPMJCR2114,
Keio University Global Research Institute (KGRI) Challenge Grant,
and JSPS Kakenhi Grant Numbers JP24K00004, JP21K00016, JP21H00467, JP23K20416, and JP21K18339.}

\clearpage
\appendix
\onecolumn

\section{Dataset Description}
\label{appendix:dataset}

\paragraph{Epistemic Syllogism.}
All formal patterns of epistemic syllogisms examined in this work can be found in Table~\ref{tab:mp-schema-epistemic}.

\begin{table*}[th]
\centering
\scalebox{0.62}{
\begin{tabular}{llll}
\cellcolor{\corC}
\deductionCC{Cat-MP}
{All B are certain to C}{A is B}{A is certain to C} &
\cellcolor{\corC}
\deductionCC{Cat-MT}
{All B are certain to C}{A is not certain to C}{A is not B} &
\cellcolor{\corC}
\deductionCC{Hyp-MP}
{If A is B, then it is certain that A is C}{A is B}{It is certain that A is C} &
\cellcolor{\corC}
\deductionCC{Hyp-MT}
{If A is B, then it is certain that A is C}{It is not certain that A is C}{A is not B}
\medskip \\
\cellcolor{\incorC}
\deductionCC{Cat-AC}
{All B are certain to C}{A is certain to C}{A is B} &
\cellcolor{\incorC}
\deductionCC{Cat-DA}
{All B are certain to C}{A is not B}{A is not certain to C}
&
\cellcolor{\incorC}
\deductionCC{Hyp-AC}
{If A is B, then it is certain that A is C}{It is certain that A is C}{A is B} &
\cellcolor{\incorC}
\deductionCC{Hyp-DA}
{If A is B, then it is certain that A is C}{A is not B}{It is not certain that A is C}
\end{tabular}
}
\caption{
  8 patterns of epistemic syllogisms.
  \textsf{P1}: Major Premise, \textsf{P2}: Minor Premise, \textsf{C}: Conclusion;
  \texttt{Cat}: Categorical,
  \texttt{Hyp}: Hypothetical;
  \texttt{MP}: Modus Ponens,
  \texttt{MT}: Modus Tollens,
  \texttt{AC}: Affirming the Consequent,
  \texttt{DA}: Denying the Antecedent.
  $B$ and $C$ represent predicates, and $A$ represents a term.
  Those in \colorbox{\corC}{blue} are \textit{valid} patterns, while those in \colorbox{\incorC}{red} are \textit{invalid} patterns.
}
\label{tab:mp-schema-epistemic}
\end{table*}

\paragraph{Expressions of Modality.}
Table~\ref{tab:normative-template} and Table~\ref{tab:epistemic-template} present examples of natural language expressions for all types of normative and epistemic modalities (and negations) examined in this work.

\begin{table}[H]
    \centering
    \scalebox{0.8}{
    \begin{tabularx}{1.24\textwidth}{@{}lX@{}}
        \toprule
        \textbf{Form} & \textbf{Example} \\
        \midrule
        \texttt{Mu} (\textit{must} $A$) & \textit{It is obligatory to $A$,
        It is mandatory to $A$,
        There is an obligation to $A$,
        You are required to $A$,
        You must A}
        \\
        \texttt{Mi} (\textit{might} $A$) & \textit{It is permissible to $A$,
        It is acceptable to $A$,
        You are permitted to $A$,
        You are allowed to $A$,
        You can choose to $A$}
        \\
        \texttt{MuNot} (\textit{must not} $A$) & \textit{It is obligatory not to $A$,
        It is mandatory not to $A$,
        There is an obligation not to $A$,
        You are required not to $A$,
        You must not A}
        \\
        \texttt{MiNot} (\textit{might not} $A$) & \textit{It is permissible not to $A$,
        It is acceptable not to $A$,
        You are permitted not to $A$,
        You are allowed not to $A$,
        You can choose not to $A$}
        \\
        \texttt{NotMu} (\textit{not required to} $A$) & \textit{It is not obligatory to $A$,
        It is not mandatory to $A$,
        There is no obligation to $A$,
        You are not required to $A$,
        It is not the case that you must $A$}
        \\
        \texttt{NotMi} (\textit{not permitted to} $A$) & \textit{It is not permissible to $A$,
        It is not acceptable to $A$,
        You are not permitted to $A$,
        You are not allowed to $A$,
        You cannot choose to $A$}
        \\
        \bottomrule
    \end{tabularx}
    }%
    \caption{6 types of normative modality (and negations) and examples of natural language expressions for each.
    \texttt{Mu}: obligation,
    \texttt{Mi}: permission,
    \texttt{MuNot}: obligation of negation,
    \texttt{MiNot}: permission of negation,
    \texttt{NotMu}: negations of obligation,
    \texttt{NotMi}: negations of permission.
    }
    \label{tab:normative-template}
\end{table}

\begin{table}[H]
    \centering
    \scalebox{0.8}{
    \begin{tabularx}{1.24\textwidth}{@{}lX@{}}
        \toprule
        \textbf{Form} & \textbf{Example} \\
        \midrule
        \texttt{Mu} ($S$ \textit{must} $P$) & \textit{It is certain that S $P$, It is necessarily the case that $S$ $P$, It is known that $S$ $P$, It is established that $S$ $P$, $S$ must have been $P$}
        \\
        \texttt{Mi} ($S$ \textit{might} $P$) & \textit{It is possible that $S$ $P$, $S$ might $P$, $S$ might have been $P$, There's a chance $S$ $P$, There is a possibility that $S$ $P$}
        \\
        \texttt{MuNot} ($S$ \textit{must not} $P$) & \textit{It is certain that $S$ not $P$, It is necessarily the case that $S$ not $P$, It is known that $S$ not $P$, It is established that $S$ not $P$, $S$ could not have been $P$}
        \\
        \texttt{MiNot} ($S$ \textit{might not} $P$) & \textit{It is possible that $S$ not $P$, $S$ might not $P$, $S$ might not have been $P$, There's a chance $S$ not $P$, There is a possibility that $S$ not $P$}
        \\
        \texttt{NotMu} (\textit{not certain that $S$ $P$}) & \textit{It is not certain that $S$ $P$, It is not necessarily the case that $S$ $P$, It is not known that $S$ $P$, It is not established that $S$ $P$, It is not the case that $S$ must have been $P$}
        \\
        \texttt{NotMi} (\textit{impossible that $S$ $P$}) & \textit{It is impossible that $S$ $P$, It is not the case that $S$ might have been $P$, There's no chance $S$ $P$, There is no possibility that $S$ $P$}
        \\
        \bottomrule
    \end{tabularx}
    }%
    \caption{6 types of epistemic modality (and negations) and examples of natural language expressions for each.
    \texttt{Mu}: epistemic necessity,
    \texttt{Mi}: epistemic possibility,
    \texttt{MuNot}: epistemic necessity of negation,
    \texttt{MiNot}: epistemic possibility of negation,
    \texttt{NotMu}: negations of epistemic necessity,
    \texttt{NotMi}: negations of epistemic possibility.
    }
    \label{tab:epistemic-template}
  \end{table}

\section{Prompt Templates and Examples}
\label{appendix:prompt}

Tables~\ref{tab:prompt-deontic} and \ref{tab:prompt-syllogistic} present examples of the prompts used for the Deontic Logic task and Syllogistic task, respectively.

{\footnotesize
\begin{longtable}{p{15.5cm}}
\toprule
Zero-Shot prompt example (Normative)\\
\midrule
\vskip-\baselineskip
\vskip-\smallskipamount
\begin{lstlisting}
Determine whether the hypothesis follows from the premise(s).
- Answer 'entailment' if the hypothesis follows from the premise(s).
- Otherwise, answer 'non-entailment'.
Respond only with 'entailment' or 'non-entailment', and nothing else.

Premise: You are not required to attend the meeting.
Hypothesis: You are permitted not to attend the meeting.
Answer:
\end{lstlisting}
\kern-\baselineskip\\ \midrule
Few-Shot prompt example (Normative, abbreviated)\\
\midrule
\vskip-\baselineskip
\vskip-\smallskipamount
\begin{lstlisting}
Determine whether the hypothesis follows from the premise(s).
- Answer 'entailment' if the hypothesis follows from the premise(s).
- Otherwise, answer 'non-entailment'.
Respond only with 'entailment' or 'non-entailment', and nothing else.

Premise: You are not required to finish homework by Friday.
Hypothesis: It is permissible not to finish homework by Friday.
Answer: entailment

[...]

Premise: It is obligatory to mail a letter.
Hypothesis: It is obligatory to mail a letter or to burn it.
Answer: non-entailment

Premise: You are not required to attend the meeting.
Hypothesis: You are permitted not to attend the meeting.
Answer:
\end{lstlisting}
\kern-\baselineskip\\ \midrule
CoT prompt example (Normative)\\
\midrule
\vskip-\baselineskip
\vskip-\smallskipamount
\begin{lstlisting}
Determine whether the hypothesis follows from the premise(s).
- Answer 'entailment' if the hypothesis follows from the premise(s).
- Otherwise, answer 'non-entailment'.
Let's think step by step. Then output only one word, 'entailment' or 'non-entailment' on the last line, immediately after a line break.

Premise: You are not required to attend the meeting.
Hypothesis: You are permitted not to attend the meeting.
Answer:
\end{lstlisting}
\kern-\baselineskip\\ \bottomrule
\caption{Prompt Examples for Deontic Logic Task.}
\label{tab:prompt-deontic}
\end{longtable}
}%

{\footnotesize
\begin{longtable}{p{15.5cm}}
\toprule
Zero-Shot prompt example (Normative)\\
\midrule
\vskip-\baselineskip
\vskip-\smallskipamount
\begin{lstlisting}
Determine whether the hypothesis follows from the premise(s).
- Answer 'entailment' if the hypothesis follows from the premise(s).
- Otherwise, answer 'non-entailment'.
Respond only with 'entailment' or 'non-entailment', and nothing else.

Premise: If you are a student, then it is obligatory to attend class. You are a student.
Hypothesis: It is obligatory to attend class.
Answer:
\end{lstlisting}
\kern-\baselineskip\\ \midrule
Few-Shot prompt example (Normative, abbreviated)\\
\midrule
\vskip-\baselineskip
\vskip-\smallskipamount
\begin{lstlisting}
Determine whether the hypothesis follows from the premise(s).
- Answer 'entailment' if the hypothesis follows from the premise(s).
- Otherwise, answer 'non-entailment'.
Respond only with 'entailment' or 'non-entailment', and nothing else.

Premise: If the traffic light is red, then Taro must stop the car. The traffic light is red.
Hypothesis: Taro must stop the car.
Answer: entailment

[...]

Premise: All first-year students at this university must submit a paper. Bob is not a first-year student at this university.
Hypothesis: Bob is not required to submit a paper.
Answer: non-entailment

Premise: If you are a student, then it is obligatory to attend class. You are a student.
Hypothesis: It is obligatory to attend class.
Answer:
\end{lstlisting}
\kern-\baselineskip\\ \midrule
CoT prompt example (Normative)\\
\midrule
\vskip-\baselineskip
\vskip-\smallskipamount
\begin{lstlisting}
Determine whether the hypothesis follows from the premise(s).
- Answer 'entailment' if the hypothesis follows from the premise(s).
- Otherwise, answer 'non-entailment'.
Let's think step by step. Then output only one word, 'entailment' or 'non-entailment' on the last line, immediately after a line break.

Premise: If you are a student, then it is obligatory to attend class. You are a student.
Hypothesis: It is obligatory to attend class.
Answer:
\end{lstlisting}
\kern-\baselineskip\\ \bottomrule
\caption{Prompt Examples for Syllogistic Task.}
\label{tab:prompt-syllogistic}
\end{longtable}
} %

\afterpage{\clearpage}

\section{Supplemental Experimental Results}
\label{appendix:experiments}

\subsection{Deontic Logic Task}

Examples of errors for each inference pattern are presented in Figures~\ref{fig:errors-mumi}, \ref{fig:errors-fc}, and \ref{fig:errors-ross}.
The complete results for the Deontic Logic task are reported in Tables~\ref{tab:sp-normative} and \ref{tab:sp-epistemic}.

\begin{figure}[H]
  \footnotesize
  \begin{minipage}{0.48\linewidth}
    \centering
    \ErrorExampleBox{Normative}{Mu-Mi}{valid}{congruent}
{You must follow the rules of the game.}
{You can choose to follow the rules of the game.}
{\ModelRes
{Few-shot}
{\ModelFailureNonEnt} %
{\ModelFailureNonEnt} %
{\ModelFailureNonEnt} %
{\ModelFailureNonEnt} %
{\ModelFailureNonEnt} %
}
  \end{minipage}
  \hfill
  \begin{minipage}{0.48\linewidth}
    \centering
    \ErrorExampleBox{Normative}{Mu-Mi}{valid}{congruent}
{You must take care of your health.}
{You can choose to take care of your health.}
{\ModelRes
{Few-shot}
{\ModelFailureNonEnt} %
{\ModelFailureNonEnt} %
{\ModelFailureNonEnt} %
{\ModelSuccessEnt} %
{\ModelSuccessEnt} %
}
  \end{minipage}
  \caption{Examples of errors for \texttt{Mu-Mi}.}
  \label{fig:errors-mumi}
\end{figure}

\begin{figure}[H]
\centering
\footnotesize
\ErrorExampleBox{Normative}{FC-Or-Intro}{invalid}{congruent}
{You may eat cake.}
{You may eat cake or cookies.}
{\ModelRes
{Zero-shot}
{\ModelFailureEnt}  %
{\ModelFailureEnt}  %
{\ModelSuccessNonEnt}  %
{\ModelFailureEnt}  %
{\ModelSuccessNonEnt}  %
}
\caption{An example of error for \texttt{FC-Or-Intro}.}
\label{fig:errors-fc}
\vspace{2em}

\ErrorExampleBox{Normative}{Ross-Or-Intro}{invalid}{incongruent}
{You must pay your taxes.}
{You must pay your taxes or evade them.}
{\ModelRes
{Zero-shot}
{\ModelSuccessNonEnt}  %
{\ModelFailureEnt}  %
{\ModelFailureEnt}  %
{\ModelFailureEnt}  %
{\ModelFailureEnt}  %
}
\caption{An example of error for \texttt{Ross-Or-Intro}.}
\label{fig:errors-ross}
\end{figure}

\begin{table}[H]
\centering
\begin{subtable}{\linewidth}
\centering
\caption{gold = \textit{entailment}.}
\scalebox{0.475}{
\begin{tabular}{l|ccc>{\columncolor{lightgray!25}}c ccc>{\columncolor{lightgray!25}}c ccc>{\columncolor{lightgray!25}}c ccc>{\columncolor{lightgray!25}}c ccc>{\columncolor{lightgray!25}}c ccc>{\columncolor{lightgray!25}}c}
\toprule
\multirow{2}{*}{\textbf{Model}}
  & \multicolumn{4}{c}{\textbf{NotMu-MiNot}}
  & \multicolumn{4}{c}{\textbf{NotMi-MuNot}}
  & \multicolumn{4}{c}{\textbf{MiNot-NotMu}}
  & \multicolumn{4}{c}{\textbf{Mu-Mi}}
  & \multicolumn{4}{c}{\textbf{NotMi-NotMu}}
  & \multicolumn{4}{c}{\textbf{FC-Or-Elim}} \\
  \cmidrule(lr){2-5}\cmidrule(lr){6-9}\cmidrule(lr){10-13}\cmidrule(lr){14-17}\cmidrule(lr){18-21}\cmidrule(lr){22-25}
  & \textbf{C} & \textbf{I} & \textbf{N} & \textbf{Avg.} & \textbf{C} & \textbf{I} & \textbf{N} & \textbf{Avg.} & \textbf{C} & \textbf{I} & \textbf{N} & \textbf{Avg.} & \textbf{C} & \textbf{I} & \textbf{N} & \textbf{Avg.} & \textbf{C} & \textbf{I} & \textbf{N} & \textbf{Avg.} & \textbf{C} & \textbf{I} & \textbf{N} & \textbf{Avg.} \\
\midrule
gpt-4o-mini &
  100.0 & 100.0 & 100.0 & 100.0 &
  100.0 & 45.0 & 85.0 & 76.67 &
  100.0 & 100.0 & 100.0 & 100.0 &
  35.0 & 70.0 & 10.0 & 38.33 &
  80.0 & 85.0 & 80.0 & 81.67 &
  25.0 & 15.0 & 100.0 & 46.67 \\
\hspace{1em}+ Few-Shot &
  100.0 & 95.0 & 100.0 & 98.33 &
  95.0 & 35.0 & 35.0 & 55.0 &
  100.0 & 100.0 & 100.0 & 100.0 &
  80.0 & 95.0 & 80.0 & 85.0 &
  100.0 & 95.0 & 95.0 & 96.67 &
  55.0 & 35.0 & 35.0 & 41.67 \\
\hspace{1em}+ CoT &
  100.0 & 90.0 & 100.0 & 96.67 &
  65.0 & 35.0 & 70.0 & 56.67 &
  100.0 & 95.0 & 100.0 & 98.33 &
  30.0 & 10.0 & 15.0 & 18.33 &
  90.0 & 100.0 & 80.0 & 90.0 &
  60.0 & 20.0 & 90.0 & 56.67 \\
gpt-4o &
  100.0 & 90.0 & 100.0 & 96.67 &
  100.0 & 45.0 & 70.0 & 71.67 &
  100.0 & 100.0 & 100.0 & 100.0 &
  70.0 & 55.0 & 45.0 & 56.67 &
  100.0 & 100.0 & 100.0 & 100.0 &
  70.0 & 10.0 & 90.0 & 56.67 \\
\hspace{1em}+ Few-Shot &
  100.0 & 90.0 & 100.0 & 96.67 &
  100.0 & 85.0 & 85.0 & 90.0 &
  100.0 & 100.0 & 100.0 & 100.0 &
  80.0 & 85.0 & 80.0 & 81.67 &
  100.0 & 100.0 & 100.0 & 100.0 &
  100.0 & 100.0 & 100.0 & 100.0 \\
\hspace{1em}+ CoT &
  100.0 & 90.0 & 100.0 & 96.67 &
  60.0 & 0.0 & 35.0 & 31.67 &
  95.0 & 90.0 & 100.0 & 95.0 &
  80.0 & 75.0 & 80.0 & 78.33 &
  80.0 & 85.0 & 70.0 & 78.33 &
  80.0 & 10.0 & 100.0 & 63.33 \\
llama-3.1-8B-In &
  100.0 & 100.0 & 100.0 & 100.0 &
  100.0 & 80.0 & 100.0 & 93.33 &
  100.0 & 100.0 & 95.0 & 98.33 &
  60.0 & 30.0 & 25.0 & 38.33 &
  85.0 & 80.0 & 95.0 & 86.67 &
  45.0 & 25.0 & 10.0 & 26.67 \\
\hspace{1em}+ Few-Shot &
  100.0 & 95.0 & 95.0 & 96.67 &
  100.0 & 70.0 & 85.0 & 85.0 &
  100.0 & 90.0 & 90.0 & 93.33 &
  45.0 & 0.0 & 0.0 & 15.0 &
  100.0 & 95.0 & 95.0 & 96.67 &
  0.0 & 0.0 & 0.0 & 0.0 \\
\hspace{1em}+ CoT &
  65.0 & 55.0 & 70.0 & 63.33 &
  75.0 & 50.0 & 55.0 & 60.0 &
  60.0 & 55.0 & 65.0 & 60.0 &
  35.0 & 20.0 & 45.0 & 33.33 &
  45.0 & 60.0 & 50.0 & 51.67 &
  50.0 & 35.0 & 45.0 & 43.33 \\
llama-3.3-70B-In &
  100.0 & 95.0 & 100.0 & 98.33 &
  100.0 & 55.0 & 100.0 & 85.0 &
  100.0 & 95.0 & 100.0 & 98.33 &
  95.0 & 95.0 & 55.0 & 81.67 &
  100.0 & 100.0 & 100.0 & 100.0 &
  0.0 & 0.0 & 0.0 & 0.0 \\
\hspace{1em}+ Few-Shot &
  100.0 & 95.0 & 100.0 & 98.33 &
  100.0 & 90.0 & 100.0 & 96.67 &
  100.0 & 100.0 & 100.0 & 100.0 &
  90.0 & 85.0 & 65.0 & 80.0 &
  100.0 & 100.0 & 100.0 & 100.0 &
  100.0 & 100.0 & 100.0 & 100.0 \\
\hspace{1em}+ CoT &
  95.0 & 80.0 & 100.0 & 91.67 &
  95.0 & 55.0 & 95.0 & 81.67 &
  90.0 & 95.0 & 95.0 & 93.33 &
  70.0 & 90.0 & 80.0 & 80.0 &
  100.0 & 100.0 & 100.0 & 100.0 &
  100.0 & 100.0 & 100.0 & 100.0 \\
phi-4 &
  100.0 & 80.0 & 100.0 & 93.33 &
  80.0 & 30.0 & 90.0 & 66.67 &
  75.0 & 75.0 & 80.0 & 76.67 &
  85.0 & 75.0 & 45.0 & 68.33 &
  95.0 & 95.0 & 90.0 & 93.33 &
  100.0 & 100.0 & 100.0 & 100.0 \\
\hspace{1em}+ Few-Shot &
  90.0 & 60.0 & 100.0 & 83.33 &
  75.0 & 10.0 & 30.0 & 38.33 &
  75.0 & 70.0 & 100.0 & 81.67 &
  85.0 & 100.0 & 90.0 & 91.67 &
  100.0 & 100.0 & 95.0 & 98.33 &
  100.0 & 100.0 & 100.0 & 100.0 \\
\hspace{1em}+ CoT &
  95.0 & 65.0 & 90.0 & 83.33 &
  65.0 & 30.0 & 50.0 & 48.33 &
  80.0 & 75.0 & 75.0 & 76.67 &
  60.0 & 65.0 & 75.0 & 66.67 &
  75.0 & 90.0 & 45.0 & 70.0 &
  100.0 & 100.0 & 100.0 & 100.0 \\
\bottomrule
\end{tabular}
}%
\end{subtable}

\begin{subtable}{\linewidth}
\centering
\caption{gold = \textit{non-entailment}.}
\scalebox{0.5}{
\begin{tabular}{l|ccc>{\columncolor{lightgray!25}}c ccc>{\columncolor{lightgray!25}}c ccc>{\columncolor{lightgray!25}}c ccc>{\columncolor{lightgray!25}}c ccc>{\columncolor{lightgray!25}}c}
\toprule
\multirow{2}{*}{\textbf{Model}}
  & \multicolumn{4}{c}{\textbf{NotMu-NotMi}}
  & \multicolumn{4}{c}{\textbf{MiNot-MuNot}}
  & \multicolumn{4}{c}{\textbf{Mi-Mu}}
  & \multicolumn{4}{c}{\textbf{FC-Or-Intro}}
  & \multicolumn{4}{c}{\textbf{Ross-Or-Intro}} \\
  \cmidrule(lr){2-5}\cmidrule(lr){6-9}\cmidrule(lr){10-13}\cmidrule(lr){14-17}\cmidrule(lr){18-21}
  & \textbf{C} & \textbf{I} & \textbf{N} & \textbf{Avg.} & \textbf{C} & \textbf{I} & \textbf{N} & \textbf{Avg.} & \textbf{C} & \textbf{I} & \textbf{N} & \textbf{Avg.} & \textbf{C} & \textbf{I} & \textbf{N} & \textbf{Avg.} & \textbf{C} & \textbf{I} & \textbf{N} & \textbf{Avg.} \\
\midrule
gpt-4o-mini &
  100.0 & 100.0 & 100.0 & 100.0 &
  100.0 & 100.0 & 100.0 & 100.0 &
  100.0 & 100.0 & 100.0 & 100.0 &
  100.0 & 75.0 & 100.0 & 91.67 &
  - & 80.0 & 20.0 & 50.0 \\
\hspace{1em}+ Few-Shot &
  100.0 & 100.0 & 100.0 & 100.0 &
  100.0 & 100.0 & 100.0 & 100.0 &
  100.0 & 100.0 & 100.0 & 100.0 &
  100.0 & 100.0 & 100.0 & 100.0 &
  - & 100.0 & 60.0 & 80.0 \\
\hspace{1em}+ CoT &
  100.0 & 100.0 & 100.0 & 100.0 &
  100.0 & 100.0 & 100.0 & 100.0 &
  100.0 & 100.0 & 100.0 & 100.0 &
  100.0 & 80.0 & 100.0 & 93.33 &
  - & 100.0 & 45.0 & 72.5 \\
gpt-4o &
  100.0 & 100.0 & 100.0 & 100.0 &
  100.0 & 100.0 & 100.0 & 100.0 &
  100.0 & 100.0 & 100.0 & 100.0 &
  100.0 & 70.0 & 100.0 & 90.0 &
  - & 25.0 & 55.0 & 40.0 \\
\hspace{1em}+ Few-Shot &
  100.0 & 100.0 & 100.0 & 100.0 &
  100.0 & 100.0 & 100.0 & 100.0 &
  100.0 & 100.0 & 100.0 & 100.0 &
  100.0 & 100.0 & 100.0 & 100.0 &
  - & 100.0 & 100.0 & 100.0 \\
\hspace{1em}+ CoT &
  100.0 & 100.0 & 100.0 & 100.0 &
  100.0 & 100.0 & 100.0 & 100.0 &
  100.0 & 100.0 & 100.0 & 100.0 &
  100.0 & 100.0 & 100.0 & 100.0 &
  - & 100.0 & 100.0 & 100.0 \\
llama-3.1-8B-In &
  70.0 & 80.0 & 15.0 & 55.0 &
  55.0 & 60.0 & 25.0 & 46.67 &
  75.0 & 85.0 & 50.0 & 70.0 &
  100.0 & 100.0 & 100.0 & 100.0 &
  - & 60.0 & 45.0 & 52.5 \\
\hspace{1em}+ Few-Shot &
  90.0 & 95.0 & 60.0 & 81.67 &
  65.0 & 90.0 & 60.0 & 71.67 &
  100.0 & 100.0 & 100.0 & 100.0 &
  100.0 & 100.0 & 100.0 & 100.0 &
  - & 65.0 & 60.0 & 62.5 \\
\hspace{1em}+ CoT &
  70.0 & 60.0 & 55.0 & 61.67 &
  20.0 & 55.0 & 25.0 & 33.33 &
  30.0 & 60.0 & 70.0 & 53.33 &
  65.0 & 65.0 & 90.0 & 73.33 &
  - & 25.0 & 30.0 & 27.5 \\
llama-3.3-70B-In &
  100.0 & 100.0 & 100.0 & 100.0 &
  60.0 & 80.0 & 70.0 & 70.0 &
  100.0 & 100.0 & 100.0 & 100.0 &
  100.0 & 100.0 & 100.0 & 100.0 &
  - & 20.0 & 0.0 & 10.0 \\
\hspace{1em}+ Few-Shot &
  100.0 & 100.0 & 100.0 & 100.0 &
  60.0 & 65.0 & 60.0 & 61.67 &
  100.0 & 100.0 & 100.0 & 100.0 &
  100.0 & 100.0 & 100.0 & 100.0 &
  - & 100.0 & 100.0 & 100.0 \\
\hspace{1em}+ CoT &
  100.0 & 100.0 & 100.0 & 100.0 &
  70.0 & 80.0 & 85.0 & 78.33 &
  100.0 & 100.0 & 100.0 & 100.0 &
  30.0 & 5.0 & 25.0 & 20.0 &
  - & 50.0 & 0.0 & 25.0 \\
phi-4 &
  100.0 & 100.0 & 100.0 & 100.0 &
  60.0 & 80.0 & 60.0 & 66.67 &
  100.0 & 100.0 & 100.0 & 100.0 &
  85.0 & 35.0 & 90.0 & 70.0 &
  - & 65.0 & 70.0 & 67.5 \\
\hspace{1em}+ Few-Shot &
  100.0 & 100.0 & 100.0 & 100.0 &
  65.0 & 100.0 & 70.0 & 78.33 &
  100.0 & 100.0 & 100.0 & 100.0 &
  75.0 & 25.0 & 10.0 & 36.67 &
  - & 100.0 & 65.0 & 82.5 \\
\hspace{1em}+ CoT &
  100.0 & 95.0 & 95.0 & 96.67 &
  85.0 & 90.0 & 70.0 & 81.67 &
  95.0 & 100.0 & 100.0 & 98.33 &
  65.0 & 35.0 & 100.0 & 66.67 &
  - & 100.0 & 50.0 & 75.0 \\
\bottomrule
\end{tabular}
}%
\end{subtable}

\caption{%
Accuracy (\%) for \textbf{Normative} Problems in Deontic Logic Task. \textbf{C} = Congruent, \textbf{I} = Incongruent, \textbf{N} = Nonsense, \textbf{Avg.} = Average.%
}
\label{tab:sp-normative}
\end{table}

\begin{table}[H]
\centering
\begin{subtable}{\linewidth}
\centering
\caption{gold = \textit{entailment}.}
\scalebox{0.55}{
\begin{tabular}{l|ccc>{\columncolor{lightgray!25}}c ccc>{\columncolor{lightgray!25}}c ccc>{\columncolor{lightgray!25}}c ccc>{\columncolor{lightgray!25}}c ccc>{\columncolor{lightgray!25}}c}
\toprule
\multirow{2}{*}{\textbf{Model}}
  & \multicolumn{4}{c}{\textbf{NotMu-MiNot}}
  & \multicolumn{4}{c}{\textbf{NotMi-MuNot}}
  & \multicolumn{4}{c}{\textbf{MiNot-NotMu}}
  & \multicolumn{4}{c}{\textbf{Mu-Mi}}
  & \multicolumn{4}{c}{\textbf{NotMi-NotMu}} \\
  \cmidrule(lr){2-5}\cmidrule(lr){6-9}\cmidrule(lr){10-13}\cmidrule(lr){14-17}\cmidrule(lr){18-21}
  & \textbf{C} & \textbf{I} & \textbf{N} & \textbf{Avg.} & \textbf{C} & \textbf{I} & \textbf{N} & \textbf{Avg.} & \textbf{C} & \textbf{I} & \textbf{N} & \textbf{Avg.} & \textbf{C} & \textbf{I} & \textbf{N} & \textbf{Avg.} & \textbf{C} & \textbf{I} & \textbf{N} & \textbf{Avg.} \\
\midrule
gpt-4o-mini &
  100.0 & 100.0 & 100.0 & 100.0 &
  85.0 & 55.0 & 80.0 & 73.33 &
  100.0 & 85.0 & 85.0 & 90.0 &
  85.0 & 15.0 & 85.0 & 61.67 &
  80.0 & 20.0 & 65.0 & 55.0 \\
\hspace{1em}+ Few-Shot &
  100.0 & 100.0 & 100.0 & 100.0 &
  45.0 & 55.0 & 80.0 & 60.0 &
  100.0 & 80.0 & 100.0 & 93.33 &
  90.0 & 35.0 & 100.0 & 75.0 &
  90.0 & 20.0 & 50.0 & 53.33 \\
\hspace{1em}+ CoT &
  100.0 & 100.0 & 100.0 & 100.0 &
  80.0 & 85.0 & 85.0 & 83.33 &
  100.0 & 90.0 & 100.0 & 96.67 &
  35.0 & 25.0 & 65.0 & 41.67 &
  55.0 & 15.0 & 35.0 & 35.0 \\
gpt-4o &
  100.0 & 100.0 & 100.0 & 100.0 &
  85.0 & 85.0 & 90.0 & 86.67 &
  100.0 & 90.0 & 100.0 & 96.67 &
  90.0 & 85.0 & 100.0 & 91.67 &
  100.0 & 30.0 & 95.0 & 75.0 \\
\hspace{1em}+ Few-Shot &
  100.0 & 90.0 & 100.0 & 96.67 &
  90.0 & 100.0 & 85.0 & 91.67 &
  100.0 & 80.0 & 100.0 & 93.33 &
  100.0 & 75.0 & 100.0 & 91.67 &
  80.0 & 30.0 & 85.0 & 65.0 \\
\hspace{1em}+ CoT &
  100.0 & 100.0 & 100.0 & 100.0 &
  85.0 & 95.0 & 85.0 & 88.33 &
  100.0 & 75.0 & 95.0 & 90.0 &
  85.0 & 90.0 & 100.0 & 91.67 &
  70.0 & 25.0 & 95.0 & 63.33 \\
llama-3.1-8B-In &
  100.0 & 100.0 & 100.0 & 100.0 &
  100.0 & 90.0 & 100.0 & 96.67 &
  100.0 & 100.0 & 100.0 & 100.0 &
  35.0 & 50.0 & 35.0 & 40.0 &
  100.0 & 65.0 & 100.0 & 88.33 \\
\hspace{1em}+ Few-Shot &
  100.0 & 100.0 & 100.0 & 100.0 &
  100.0 & 80.0 & 95.0 & 91.67 &
  100.0 & 100.0 & 90.0 & 96.67 &
  80.0 & 65.0 & 85.0 & 76.67 &
  95.0 & 65.0 & 85.0 & 81.67 \\
\hspace{1em}+ CoT &
  55.0 & 75.0 & 60.0 & 63.33 &
  75.0 & 60.0 & 95.0 & 76.67 &
  55.0 & 35.0 & 60.0 & 50.0 &
  5.0 & 0.0 & 5.0 & 3.33 &
  55.0 & 15.0 & 70.0 & 46.67 \\
llama-3.3-70B-In &
  100.0 & 100.0 & 100.0 & 100.0 &
  95.0 & 70.0 & 85.0 & 83.33 &
  100.0 & 95.0 & 100.0 & 98.33 &
  100.0 & 90.0 & 100.0 & 96.67 &
  80.0 & 20.0 & 80.0 & 60.0 \\
\hspace{1em}+ Few-Shot &
  100.0 & 100.0 & 100.0 & 100.0 &
  90.0 & 90.0 & 100.0 & 93.33 &
  100.0 & 80.0 & 100.0 & 93.33 &
  95.0 & 80.0 & 70.0 & 81.67 &
  85.0 & 20.0 & 80.0 & 61.67 \\
\hspace{1em}+ CoT &
  100.0 & 100.0 & 100.0 & 100.0 &
  100.0 & 95.0 & 100.0 & 98.33 &
  100.0 & 100.0 & 100.0 & 100.0 &
  100.0 & 95.0 & 100.0 & 98.33 &
  100.0 & 30.0 & 100.0 & 76.67 \\
phi-4 &
  100.0 & 100.0 & 100.0 & 100.0 &
  100.0 & 95.0 & 90.0 & 95.0 &
  100.0 & 95.0 & 95.0 & 96.67 &
  100.0 & 80.0 & 100.0 & 93.33 &
  90.0 & 25.0 & 85.0 & 66.67 \\
\hspace{1em}+ Few-Shot &
  100.0 & 100.0 & 100.0 & 100.0 &
  30.0 & 55.0 & 55.0 & 46.67 &
  85.0 & 70.0 & 90.0 & 81.67 &
  100.0 & 70.0 & 100.0 & 90.0 &
  75.0 & 20.0 & 85.0 & 60.0 \\
\hspace{1em}+ CoT &
  95.0 & 100.0 & 100.0 & 98.33 &
  95.0 & 90.0 & 95.0 & 93.33 &
  100.0 & 70.0 & 100.0 & 90.0 &
  80.0 & 70.0 & 80.0 & 76.67 &
  70.0 & 15.0 & 60.0 & 48.33 \\
\bottomrule
\end{tabular}
}%
\end{subtable}

\begin{subtable}{\linewidth}
\centering
\caption{gold = \textit{non-entailment}.}
\scalebox{0.6}{
\begin{tabular}{l|ccc>{\columncolor{lightgray!25}}c ccc>{\columncolor{lightgray!25}}c ccc>{\columncolor{lightgray!25}}c}
\toprule
\multirow{2}{*}{\textbf{Model}}
  & \multicolumn{4}{c}{\textbf{NotMu-NotMi}}
  & \multicolumn{4}{c}{\textbf{MiNot-MuNot}}
  & \multicolumn{4}{c}{\textbf{Mi-Mu}} \\
  \cmidrule(lr){2-5}\cmidrule(lr){6-9}\cmidrule(lr){10-13}
  & \textbf{C} & \textbf{I} & \textbf{N} & \textbf{Avg.} & \textbf{C} & \textbf{I} & \textbf{N} & \textbf{Avg.} & \textbf{C} & \textbf{I} & \textbf{N} & \textbf{Avg.} \\
\midrule
gpt-4o-mini &
  95.0 & 100.0 & 100.0 & 98.33 &
  100.0 & 100.0 & 100.0 & 100.0 &
  100.0 & 100.0 & 100.0 & 100.0 \\
\hspace{1em}+ Few-Shot &
  100.0 & 100.0 & 100.0 & 100.0 &
  100.0 & 100.0 & 100.0 & 100.0 &
  100.0 & 100.0 & 100.0 & 100.0 \\
\hspace{1em}+ CoT &
  100.0 & 100.0 & 100.0 & 100.0 &
  100.0 & 100.0 & 100.0 & 100.0 &
  100.0 & 100.0 & 100.0 & 100.0 \\
gpt-4o &
  100.0 & 100.0 & 100.0 & 100.0 &
  100.0 & 100.0 & 100.0 & 100.0 &
  100.0 & 100.0 & 100.0 & 100.0 \\
\hspace{1em}+ Few-Shot &
  100.0 & 100.0 & 100.0 & 100.0 &
  100.0 & 100.0 & 100.0 & 100.0 &
  100.0 & 100.0 & 100.0 & 100.0 \\
\hspace{1em}+ CoT &
  100.0 & 100.0 & 100.0 & 100.0 &
  100.0 & 100.0 & 100.0 & 100.0 &
  100.0 & 100.0 & 100.0 & 100.0 \\
llama-3.1-8B-In &
  70.0 & 55.0 & 0.0 & 41.67 &
  75.0 & 70.0 & 40.0 & 61.67 &
  85.0 & 90.0 & 65.0 & 80.0 \\
\hspace{1em}+ Few-Shot &
  30.0 & 95.0 & 20.0 & 48.33 &
  100.0 & 90.0 & 85.0 & 91.67 &
  100.0 & 100.0 & 100.0 & 100.0 \\
\hspace{1em}+ CoT &
  95.0 & 100.0 & 100.0 & 98.33 &
  100.0 & 100.0 & 100.0 & 100.0 &
  100.0 & 100.0 & 100.0 & 100.0 \\
llama-3.3-70B-In &
  85.0 & 100.0 & 100.0 & 95.0 &
  100.0 & 100.0 & 100.0 & 100.0 &
  100.0 & 100.0 & 100.0 & 100.0 \\
\hspace{1em}+ Few-Shot &
  100.0 & 100.0 & 100.0 & 100.0 &
  100.0 & 100.0 & 100.0 & 100.0 &
  100.0 & 100.0 & 100.0 & 100.0 \\
\hspace{1em}+ CoT &
  75.0 & 100.0 & 100.0 & 91.67 &
  100.0 & 100.0 & 100.0 & 100.0 &
  100.0 & 100.0 & 100.0 & 100.0 \\
phi-4 &
  75.0 & 100.0 & 80.0 & 85.0 &
  100.0 & 100.0 & 100.0 & 100.0 &
  100.0 & 100.0 & 100.0 & 100.0 \\
\hspace{1em}+ Few-Shot &
  100.0 & 100.0 & 100.0 & 100.0 &
  100.0 & 100.0 & 100.0 & 100.0 &
  100.0 & 100.0 & 100.0 & 100.0 \\
\hspace{1em}+ CoT &
  100.0 & 100.0 & 100.0 & 100.0 &
  100.0 & 100.0 & 100.0 & 100.0 &
  100.0 & 100.0 & 100.0 & 100.0 \\
\bottomrule
\end{tabular}
}%
\end{subtable}

\caption{%
Accuracy (\%) for \textbf{Epistemic} Problems in Deontic Logic Task. \textbf{C} = Congruent, \textbf{I} = Incongruent, \textbf{N} = Nonsense, \textbf{Avg.} = Average.%
}
\label{tab:sp-epistemic}
\end{table}

\subsection{Syllogistic Task}

Examples of errors for each inference pattern are presented in Figures~\ref{fig:errors-cat-da}, \ref{fig:errors-hyp-da}, \ref{fig:errors-cat-mt}, and \ref{fig:errors-hyp-mt}.
The complete results for the Syllogistic task are reported in Tables~\ref{tab:mp-normative} and \ref{tab:mp-epistemic}.

\begin{figure}[H]
\centering
\footnotesize
  \begin{minipage}{0.48\linewidth}
\ErrorExampleSylloBox{Normative}{Cat-DA}{invalid}{congruent}
{All customers must pay for their purchases.}
{A window shopper is not a customer.}
{There is no obligation for a window shopper to pay for their purchases.}
{\ModelRes
{Few-shot}
{\ModelFailureEnt}  %
{\ModelFailureEnt}  %
{\ModelFailureEnt}  %
{\ModelFailureEnt}  %
{\ModelFailureEnt}  %
}
\end{minipage}
\hfill
  \begin{minipage}{0.48\linewidth}
\ErrorExampleSylloBox{Epistemic}{Cat-DA}{invalid}{incongruent}
{All flowers necessarily have petals.}
{Sunflowers are not flowers.}
{Sunflowers do not necessarily have petals.}
{\ModelRes
{Few-shot}
{\ModelFailureEnt}  %
{\ModelFailureEnt}  %
{\ModelFailureEnt}  %
{\ModelFailureEnt}  %
{\ModelSuccessNonEnt}  %
}
\end{minipage}
\caption{Examples of errors for \texttt{Cat-DA}.}
\label{fig:errors-cat-da}
\end{figure}

\begin{figure}[H]
\centering
\footnotesize
\begin{minipage}{0.49\linewidth}
\ErrorExampleSylloBox{Normative}{Hyp-DA}{invalid}{congruent}
{If you want to drive a car, then it is obligatory to have a driver's license.}
{You do not want to drive a car.}
{It is not obligatory to have a driver's license.}
{\ModelRes
{Few-shot}
{\ModelSuccessNonEnt}  %
{\ModelSuccessNonEnt}  %
{\ModelFailureEnt}  %
{\ModelFailureEnt}  %
{\ModelSuccessNonEnt}  %
}
\end{minipage}
\hfill
\begin{minipage}{0.49\linewidth}
\ErrorExampleSylloBox{Epistemic}{Hyp-DA}{invalid}{congruent}
{If the expedition locates the ancient ruins, then it is established that they will make a significant archaeological discovery.}
{The expedition did not locate the ancient ruins.}
{It is not established that the expedition will make a significant archaeological discovery.}
{\ModelRes
{Few-shot}
{\ModelFailureEnt}  %
{\ModelSuccessNonEnt}  %
{\ModelFailureEnt}  %
{\ModelFailureEnt}  %
{\ModelSuccessNonEnt}  %
}
\end{minipage}
\caption{Examples of errors for \texttt{Hyp-DA}.}
\label{fig:errors-hyp-da}
\end{figure}

\begin{figure}[H]
\footnotesize
\ErrorExampleSylloBox{Normative}{Cat-MT}{valid}{congruent}
{All teachers must be qualified.}
{It is not the case that she must be qualified.}
{She is not a teacher.}
{\ModelRes
{Few-shot}
{\ModelSuccessEnt}  %
{\ModelSuccessEnt}  %
{\ModelFailureNonEnt}  %
{\ModelSuccessEnt}  %
{\ModelFailureNonEnt}  %
}
\caption{An example of an error for \texttt{Cat-MT}.}
\label{fig:errors-cat-mt}
\vspace{2em}

\ErrorExampleSylloBox{Normative}{Hyp-MT}{valid}{congruent}
{If you are invited to a wedding, then it is customary to bring a gift.}
{It is not the case that you are required to bring a gift.}
{You are not invited to a wedding.}
{\ModelRes
{Few-shot}
{\ModelSuccessEnt}  %
{\ModelSuccessEnt}  %
{\ModelFailureNonEnt}  %
{\ModelSuccessEnt}  %
{\ModelFailureNonEnt}  %
}
\caption{An example of an error for \texttt{Hyp-MT}.}
\label{fig:errors-hyp-mt}
\end{figure}

\begin{table}[H]
\centering
\begin{subtable}{\linewidth}
\centering
\caption{gold = \textit{entailment}.}
\scalebox{0.6}{
\begin{tabular}{l|ccc>{\columncolor{lightgray!25}}c ccc>{\columncolor{lightgray!25}}c ccc>{\columncolor{lightgray!25}}c ccc>{\columncolor{lightgray!25}}c}
\toprule
\multirow{2}{*}{\textbf{Model}}
  & \multicolumn{4}{c}{\textbf{Cat-MP}}
  & \multicolumn{4}{c}{\textbf{Cat-MT}}
  & \multicolumn{4}{c}{\textbf{Hyp-MP}}
  & \multicolumn{4}{c}{\textbf{Hyp-MT}} \\
  \cmidrule(lr){2-5}
  \cmidrule(lr){6-9}
  \cmidrule(lr){10-13}
  \cmidrule(lr){14-17}
  & \textbf{C} & \textbf{I} & \textbf{N} & \textbf{Avg.} &
  \textbf{C} & \textbf{I} & \textbf{N} & \textbf{Avg.} &
  \textbf{C} & \textbf{I} & \textbf{N} & \textbf{Avg.} &
  \textbf{C} & \textbf{I} & \textbf{N} & \textbf{Avg.} \\
\midrule
gpt-4o-mini &
  100.0 & 100.0 & 100.0 & 100.0 &
  90.0 & 20.0 & 25.0 & 45.0 &
  100.0 & 100.0 & 100.0 & 100.0 &
  50.0 & 30.0 & 100.0 & 60.0 \\
\hspace{1em}+ Few-Shot &
  100.0 & 100.0 & 100.0 & 100.0 &
  100.0 & 65.0 & 45.0 & 70.0 &
  100.0 & 100.0 & 100.0 & 100.0 &
  70.0 & 30.0 & 100.0 & 66.67 \\
\hspace{1em}+ CoT &
  100.0 & 100.0 & 100.0 & 100.0 &
  85.0 & 35.0 & 25.0 & 48.33 &
  100.0 & 100.0 & 100.0 & 100.0 &
  50.0 & 50.0 & 100.0 & 66.67 \\
gpt-4o &
  100.0 & 100.0 & 100.0 & 100.0 &
  90.0 & 85.0 & 25.0 & 66.67 &
  100.0 & 100.0 & 100.0 & 100.0 &
  85.0 & 90.0 & 100.0 & 91.67 \\
\hspace{1em}+ Few-Shot &
  100.0 & 100.0 & 100.0 & 100.0 &
  100.0 & 100.0 & 80.0 & 93.33 &
  100.0 & 100.0 & 100.0 & 100.0 &
  100.0 & 60.0 & 100.0 & 86.67 \\
\hspace{1em}+ CoT &
  100.0 & 100.0 & 100.0 & 100.0 &
  100.0 & 100.0 & 25.0 & 75.0 &
  100.0 & 100.0 & 100.0 & 100.0 &
  60.0 & 70.0 & 100.0 & 76.67 \\
llama-3.1-8B-In &
  100.0 & 100.0 & 100.0 & 100.0 &
  90.0 & 95.0 & 25.0 & 70.0 &
  100.0 & 95.0 & 100.0 & 98.33 &
  80.0 & 50.0 & 100.0 & 76.67 \\
\hspace{1em}+ Few-Shot &
  100.0 & 100.0 & 60.0 & 86.67 &
  65.0 & 85.0 & 15.0 & 55.0 &
  100.0 & 100.0 & 100.0 & 100.0 &
  70.0 & 95.0 & 100.0 & 88.33 \\
\hspace{1em}+ CoT &
  100.0 & 95.0 & 90.0 & 95.0 &
  65.0 & 35.0 & 15.0 & 38.33 &
  100.0 & 100.0 & 100.0 & 100.0 &
  25.0 & 25.0 & 80.0 & 43.33 \\
llama-3.3-70B-In &
  100.0 & 100.0 & 100.0 & 100.0 &
  90.0 & 60.0 & 20.0 & 56.67 &
  100.0 & 100.0 & 100.0 & 100.0 &
  50.0 & 55.0 & 100.0 & 68.33 \\
\hspace{1em}+ Few-Shot &
  100.0 & 100.0 & 100.0 & 100.0 &
  100.0 & 100.0 & 30.0 & 76.67 &
  100.0 & 100.0 & 100.0 & 100.0 &
  95.0 & 55.0 & 100.0 & 83.33 \\
\hspace{1em}+ CoT &
  100.0 & 95.0 & 100.0 & 98.33 &
  85.0 & 90.0 & 30.0 & 68.33 &
  100.0 & 100.0 & 100.0 & 100.0 &
  50.0 & 50.0 & 95.0 & 65.0 \\
phi-4 &
  100.0 & 100.0 & 100.0 & 100.0 &
  100.0 & 25.0 & 30.0 & 51.67 &
  100.0 & 100.0 & 100.0 & 100.0 &
  45.0 & 35.0 & 100.0 & 60.0 \\
\hspace{1em}+ Few-Shot &
  75.0 & 95.0 & 20.0 & 63.33 &
  90.0 & 10.0 & 5.0 & 35.0 &
  65.0 & 95.0 & 100.0 & 86.67 &
  30.0 & 10.0 & 85.0 & 41.67 \\
\hspace{1em}+ CoT &
  100.0 & 100.0 & 95.0 & 98.33 &
  100.0 & 90.0 & 25.0 & 71.67 &
  100.0 & 100.0 & 100.0 & 100.0 &
  60.0 & 75.0 & 100.0 & 78.33 \\
\bottomrule
\end{tabular}
}%
\end{subtable}

\bigskip

\begin{subtable}{\linewidth}
\centering
\caption{gold = \textit{non-entailment}.}
\scalebox{0.6}{
\begin{tabular}{l|ccc>{\columncolor{lightgray!25}}c ccc>{\columncolor{lightgray!25}}c ccc>{\columncolor{lightgray!25}}c ccc>{\columncolor{lightgray!25}}c}
\toprule
\multirow{2}{*}{\textbf{Model}}
  & \multicolumn{4}{c}{\textbf{Cat-AC}}
  & \multicolumn{4}{c}{\textbf{Cat-DA}}
  & \multicolumn{4}{c}{\textbf{Hyp-AC}}
  & \multicolumn{4}{c}{\textbf{Hyp-DA}} \\
  \cmidrule(lr){2-5}
  \cmidrule(lr){6-9}
  \cmidrule(lr){10-13}
  \cmidrule(lr){14-17}
  & \textbf{C} & \textbf{I} & \textbf{N} & \textbf{Avg.} &
  \textbf{C} & \textbf{I} & \textbf{N} & \textbf{Avg.} &
  \textbf{C} & \textbf{I} & \textbf{N} & \textbf{Avg.} &
  \textbf{C} & \textbf{I} & \textbf{N} & \textbf{Avg.} \\
\midrule
gpt-4o-mini &
  100.0 & 100.0 & 100.0 & 100.0 &
  100.0 & 100.0 & 100.0 & 100.0 &
  55.0 & 75.0 & 15.0 & 48.33 &
  55.0 & 75.0 & 15.0 & 48.33 \\
\hspace{1em}+ Few-Shot &
  100.0 & 100.0 & 100.0 & 100.0 &
  100.0 & 100.0 & 100.0 & 100.0 &
  100.0 & 100.0 & 100.0 & 100.0 &
  100.0 & 100.0 & 100.0 & 100.0 \\
\hspace{1em}+ CoT &
  100.0 & 100.0 & 100.0 & 100.0 &
  100.0 & 100.0 & 100.0 & 100.0 &
  85.0 & 90.0 & 100.0 & 91.67 &
  85.0 & 90.0 & 100.0 & 91.67 \\
gpt-4o &
  100.0 & 100.0 & 100.0 & 100.0 &
  100.0 & 100.0 & 100.0 & 100.0 &
  75.0 & 85.0 & 60.0 & 73.33 &
  75.0 & 85.0 & 60.0 & 73.33 \\
\hspace{1em}+ Few-Shot &
  100.0 & 100.0 & 100.0 & 100.0 &
  100.0 & 100.0 & 100.0 & 100.0 &
  100.0 & 95.0 & 100.0 & 98.33 &
  100.0 & 95.0 & 100.0 & 98.33 \\
\hspace{1em}+ CoT &
  100.0 & 100.0 & 100.0 & 100.0 &
  100.0 & 100.0 & 100.0 & 100.0 &
  100.0 & 100.0 & 100.0 & 100.0 &
  100.0 & 100.0 & 100.0 & 100.0 \\
llama-3.1-8B-In &
  35.0 & 95.0 & 5.0 & 45.0 &
  35.0 & 95.0 & 5.0 & 45.0 &
  25.0 & 30.0 & 0.0 & 18.33 &
  25.0 & 30.0 & 0.0 & 18.33 \\
\hspace{1em}+ Few-Shot &
  0.0 & 100.0 & 0.0 & 33.33 &
  0.0 & 100.0 & 0.0 & 33.33 &
  0.0 & 5.0 & 0.0 & 1.67 &
  0.0 & 5.0 & 0.0 & 1.67 \\
\hspace{1em}+ CoT &
  50.0 & 100.0 & 50.0 & 66.67 &
  50.0 & 100.0 & 50.0 & 66.67 &
  50.0 & 55.0 & 10.0 & 38.33 &
  50.0 & 55.0 & 10.0 & 38.33 \\
llama-3.3-70B-In &
  100.0 & 100.0 & 100.0 & 100.0 &
  100.0 & 100.0 & 100.0 & 100.0 &
  25.0 & 50.0 & 45.0 & 40.0 &
  25.0 & 50.0 & 45.0 & 40.0 \\
\hspace{1em}+ Few-Shot &
  100.0 & 100.0 & 100.0 & 100.0 &
  100.0 & 100.0 & 100.0 & 100.0 &
  50.0 & 55.0 & 100.0 & 68.33 &
  50.0 & 55.0 & 100.0 & 68.33 \\
\hspace{1em}+ CoT &
  100.0 & 100.0 & 100.0 & 100.0 &
  100.0 & 100.0 & 100.0 & 100.0 &
  40.0 & 30.0 & 0.0 & 23.33 &
  40.0 & 30.0 & 0.0 & 23.33 \\
phi-4 &
  100.0 & 100.0 & 100.0 & 100.0 &
  100.0 & 100.0 & 100.0 & 100.0 &
  70.0 & 75.0 & 100.0 & 81.67 &
  70.0 & 75.0 & 100.0 & 81.67 \\
\hspace{1em}+ Few-Shot &
  80.0 & 20.0 & 80.0 & 60.0 &
  80.0 & 20.0 & 80.0 & 60.0 &
  100.0 & 95.0 & 100.0 & 98.33 &
  100.0 & 95.0 & 100.0 & 98.33 \\
\hspace{1em}+ CoT &
  100.0 & 100.0 & 100.0 & 100.0 &
  100.0 & 100.0 & 100.0 & 100.0 &
  80.0 & 85.0 & 95.0 & 86.67 &
  80.0 & 85.0 & 95.0 & 86.67 \\
\bottomrule
\end{tabular}
}%
\end{subtable}

\caption{%
Accuracy (\%) for \textbf{Normative} Problems in Syllogistic Task. \textbf{C} = Congruent, \textbf{I} = Incongruent, \textbf{N} = Nonsense, \textbf{Avg.} = Average.%
}
\label{tab:mp-normative}
\end{table}

\begin{table}[H]
\centering
\begin{subtable}{\linewidth}
\centering
\caption{gold = \textit{entailment}.}
\scalebox{0.6}{
\begin{tabular}{l|ccc>{\columncolor{lightgray!25}}c ccc>{\columncolor{lightgray!25}}c ccc>{\columncolor{lightgray!25}}c ccc>{\columncolor{lightgray!25}}c}
\toprule
\multirow{2}{*}{\textbf{Model}}
  & \multicolumn{4}{c}{\textbf{Cat-MP}}
  & \multicolumn{4}{c}{\textbf{Cat-MT}}
  & \multicolumn{4}{c}{\textbf{Hyp-MP}}
  & \multicolumn{4}{c}{\textbf{Hyp-MT}} \\
  \cmidrule(lr){2-5}
  \cmidrule(lr){6-9}
  \cmidrule(lr){10-13}
  \cmidrule(lr){14-17}
  & \textbf{C} & \textbf{I} & \textbf{N} & \textbf{Avg.} &
  \textbf{C} & \textbf{I} & \textbf{N} & \textbf{Avg.} &
  \textbf{C} & \textbf{I} & \textbf{N} & \textbf{Avg.} &
  \textbf{C} & \textbf{I} & \textbf{N} & \textbf{Avg.} \\
\midrule
gpt-4o-mini &
  100.0 & 90.0 & 100.0 & 96.67 &
  70.0 & 40.0 & 100.0 & 70.0 &
  100.0 & 75.0 & 100.0 & 91.67 &
  10.0 & 60.0 & 10.0 & 26.67 \\
\hspace{1em}+ Few-Shot &
  100.0 & 75.0 & 100.0 & 91.67 &
  80.0 & 65.0 & 95.0 & 80.0 &
  100.0 & 70.0 & 100.0 & 90.0 &
  10.0 & 30.0 & 5.0 & 15.0 \\
\hspace{1em}+ CoT &
  100.0 & 100.0 & 100.0 & 100.0 &
  90.0 & 45.0 & 85.0 & 73.33 &
  100.0 & 85.0 & 100.0 & 95.0 &
  5.0 & 30.0 & 0.0 & 11.67 \\
gpt-4o &
  100.0 & 100.0 & 100.0 & 100.0 &
  95.0 & 100.0 & 100.0 & 98.33 &
  100.0 & 95.0 & 100.0 & 98.33 &
  25.0 & 80.0 & 5.0 & 36.67 \\
\hspace{1em}+ Few-Shot &
  100.0 & 100.0 & 100.0 & 100.0 &
  90.0 & 95.0 & 90.0 & 91.67 &
  100.0 & 95.0 & 100.0 & 98.33 &
  40.0 & 15.0 & 45.0 & 33.33 \\
\hspace{1em}+ CoT &
  100.0 & 100.0 & 95.0 & 98.33 &
  75.0 & 80.0 & 85.0 & 80.0 &
  100.0 & 85.0 & 100.0 & 95.0 &
  0.0 & 25.0 & 0.0 & 8.33 \\
llama-3.1-8B-In &
  100.0 & 100.0 & 100.0 & 100.0 &
  100.0 & 100.0 & 100.0 & 100.0 &
  100.0 & 75.0 & 100.0 & 91.67 &
  100.0 & 85.0 & 85.0 & 90.0 \\
\hspace{1em}+ Few-Shot &
  100.0 & 95.0 & 100.0 & 98.33 &
  100.0 & 90.0 & 100.0 & 96.67 &
  100.0 & 35.0 & 100.0 & 78.33 &
  85.0 & 90.0 & 90.0 & 88.33 \\
\hspace{1em}+ CoT &
  100.0 & 100.0 & 95.0 & 98.33 &
  70.0 & 40.0 & 50.0 & 53.33 &
  100.0 & 70.0 & 100.0 & 90.0 &
  35.0 & 50.0 & 25.0 & 36.67 \\
llama-3.3-70B-In &
  100.0 & 95.0 & 100.0 & 98.33 &
  95.0 & 45.0 & 100.0 & 80.0 &
  100.0 & 80.0 & 100.0 & 93.33 &
  0.0 & 15.0 & 5.0 & 6.67 \\
\hspace{1em}+ Few-Shot &
  100.0 & 95.0 & 100.0 & 98.33 &
  95.0 & 45.0 & 100.0 & 80.0 &
  100.0 & 80.0 & 100.0 & 93.33 &
  0.0 & 15.0 & 5.0 & 6.67 \\
\hspace{1em}+ CoT &
  100.0 & 95.0 & 100.0 & 98.33 &
  95.0 & 45.0 & 100.0 & 80.0 &
  100.0 & 80.0 & 100.0 & 93.33 &
  0.0 & 15.0 & 5.0 & 6.67 \\
phi-4 &
  100.0 & 95.0 & 100.0 & 98.33 &
  100.0 & 60.0 & 100.0 & 86.67 &
  100.0 & 95.0 & 100.0 & 98.33 &
  5.0 & 70.0 & 15.0 & 30.0 \\
\hspace{1em}+ Few-Shot &
  100.0 & 90.0 & 100.0 & 96.67 &
  100.0 & 70.0 & 100.0 & 90.0 &
  100.0 & 50.0 & 100.0 & 83.33 &
  0.0 & 10.0 & 0.0 & 3.33 \\
\hspace{1em}+ CoT &
  100.0 & 100.0 & 100.0 & 100.0 &
  100.0 & 85.0 & 100.0 & 95.0 &
  100.0 & 100.0 & 100.0 & 100.0 &
  20.0 & 50.0 & 5.0 & 25.0 \\
\bottomrule
\end{tabular}
}%
\end{subtable}

\bigskip

\begin{subtable}{\linewidth}
\centering
\caption{gold = \textit{non-entailment}.}
\scalebox{0.6}{
\begin{tabular}{l|ccc>{\columncolor{lightgray!25}}c ccc>{\columncolor{lightgray!25}}c ccc>{\columncolor{lightgray!25}}c ccc>{\columncolor{lightgray!25}}c}
\toprule
\multirow{2}{*}{\textbf{Model}}
  & \multicolumn{4}{c}{\textbf{Cat-AC}}
  & \multicolumn{4}{c}{\textbf{Cat-DA}}
  & \multicolumn{4}{c}{\textbf{Hyp-AC}}
  & \multicolumn{4}{c}{\textbf{Hyp-DA}} \\
  \cmidrule(lr){2-5}
  \cmidrule(lr){6-9}
  \cmidrule(lr){10-13}
  \cmidrule(lr){14-17}
  & \textbf{C} & \textbf{I} & \textbf{N} & \textbf{Avg.} &
  \textbf{C} & \textbf{I} & \textbf{N} & \textbf{Avg.} &
  \textbf{C} & \textbf{I} & \textbf{N} & \textbf{Avg.} &
  \textbf{C} & \textbf{I} & \textbf{N} & \textbf{Avg.} \\
\midrule
gpt-4o-mini &
  70.0 & 75.0 & 100.0 & 81.67 &
  70.0 & 75.0 & 100.0 & 81.67 &
  5.0 & 10.0 & 20.0 & 11.67 &
  5.0 & 10.0 & 20.0 & 11.67 \\
\hspace{1em}+ Few-Shot &
  90.0 & 100.0 & 95.0 & 95.0 &
  90.0 & 100.0 & 95.0 & 95.0 &
  90.0 & 75.0 & 100.0 & 88.33 &
  90.0 & 75.0 & 100.0 & 88.33 \\
\hspace{1em}+ CoT &
  100.0 & 90.0 & 100.0 & 96.67 &
  100.0 & 90.0 & 100.0 & 96.67 &
  40.0 & 90.0 & 70.0 & 66.67 &
  40.0 & 90.0 & 70.0 & 66.67 \\
gpt-4o &
  95.0 & 55.0 & 85.0 & 78.33 &
  95.0 & 55.0 & 85.0 & 78.33 &
  5.0 & 35.0 & 50.0 & 30.0 &
  5.0 & 35.0 & 50.0 & 30.0 \\
\hspace{1em}+ Few-Shot &
  100.0 & 95.0 & 100.0 & 98.33 &
  100.0 & 95.0 & 100.0 & 98.33 &
  95.0 & 95.0 & 95.0 & 95.0 &
  95.0 & 95.0 & 95.0 & 95.0 \\
\hspace{1em}+ CoT &
  100.0 & 95.0 & 100.0 & 98.33 &
  100.0 & 95.0 & 100.0 & 98.33 &
  80.0 & 75.0 & 100.0 & 85.0 &
  80.0 & 75.0 & 100.0 & 85.0 \\
llama-3.1-8B-In &
  0.0 & 45.0 & 45.0 & 30.0 &
  0.0 & 45.0 & 45.0 & 30.0 &
  5.0 & 5.0 & 0.0 & 3.33 &
  5.0 & 5.0 & 0.0 & 3.33 \\
\hspace{1em}+ Few-Shot &
  0.0 & 40.0 & 15.0 & 18.33 &
  0.0 & 40.0 & 15.0 & 18.33 &
  0.0 & 5.0 & 0.0 & 1.67 &
  0.0 & 5.0 & 0.0 & 1.67 \\
\hspace{1em}+ CoT &
  15.0 & 70.0 & 50.0 & 45.0 &
  15.0 & 70.0 & 50.0 & 45.0 &
  20.0 & 30.0 & 40.0 & 30.0 &
  20.0 & 30.0 & 40.0 & 30.0 \\
llama-3.3-70B-In &
  70.0 & 85.0 & 95.0 & 83.33 &
  70.0 & 85.0 & 95.0 & 83.33 &
  5.0 & 20.0 & 30.0 & 18.33 &
  5.0 & 20.0 & 30.0 & 18.33 \\
\hspace{1em}+ Few-Shot &
  70.0 & 85.0 & 95.0 & 83.33 &
  70.0 & 85.0 & 95.0 & 83.33 &
  5.0 & 20.0 & 30.0 & 18.33 &
  5.0 & 20.0 & 30.0 & 18.33 \\
\hspace{1em}+ CoT &
  70.0 & 85.0 & 95.0 & 83.33 &
  70.0 & 85.0 & 95.0 & 83.33 &
  5.0 & 20.0 & 30.0 & 18.33 &
  5.0 & 20.0 & 30.0 & 18.33 \\
phi-4 &
  95.0 & 75.0 & 100.0 & 90.0 &
  95.0 & 75.0 & 100.0 & 90.0 &
  30.0 & 90.0 & 35.0 & 51.67 &
  30.0 & 90.0 & 35.0 & 51.67 \\
\hspace{1em}+ Few-Shot &
  95.0 & 85.0 & 100.0 & 93.33 &
  95.0 & 85.0 & 100.0 & 93.33 &
  75.0 & 95.0 & 50.0 & 73.33 &
  75.0 & 95.0 & 50.0 & 73.33 \\
\hspace{1em}+ CoT &
  45.0 & 50.0 & 80.0 & 58.33 &
  45.0 & 50.0 & 80.0 & 58.33 &
  55.0 & 60.0 & 50.0 & 55.0 &
  55.0 & 60.0 & 50.0 & 55.0 \\
\bottomrule
\end{tabular}
}%
\end{subtable}

\caption{%
Accuracy (\%) for \textbf{Epistemic} Problems in Syllogistic Task. \textbf{C} = Congruent, \textbf{I} = Incongruent, \textbf{N} = Nonsense, \textbf{Avg.} = Average.%
}
\label{tab:mp-epistemic}
\end{table}

\end{document}